\documentclass[sigconf,screen,anonymous=False]{acmart}
\renewcommand\footnotetextcopyrightpermission[1]{} % removes footnote with conference information in first column
\makeatletter
\let\@authorsaddresses\@empty
\makeatother
\AtBeginDocument{%
  \providecommand\BibTeX{{%
    \normalfont B\kern-0.5em{\scshape i\kern-0.25em b}\kern-0.8em\TeX}}}

\setcopyright{acmcopyright}
\copyrightyear{2022}
\acmYear{2022}
\acmDOI{10.1145/1122445.1122456}

\usepackage{subfigure}
\usepackage{caption}
\usepackage{multirow}
\usepackage{enumitem}

\begin{document}

%%
%% The ``title'' command has an optional parameter,
%% allowing the author to define a ``short title'' to be used in page headers.

% \title{Towards Better and Robust Generation Framework for Recommendation Explanation: Avoiding Exaggerating Emotion Bias}

%\title{Towards Generating Robust and Fair Emotion-aware Explanation in Explainable Recommender Systems}

\title{Towards Generating Robust, Fair, and Emotion-Aware Explanations for Recommender Systems}

%%
%% The ``author'' command and its associated commands are used to define
%% ``authornote'' and ``authornotemark'' commands
    \author{Bingbing Wen}
    \affiliation{%
     \institution{\small{University of Washington}}
     \city{\small{Seattle}}
      \state{\small{WA}}
      \country{\small{US}}
    }\email{bingbw@uw.edu}
    
    \author{Yunhe Feng}
    \affiliation{%
     \institution{\small{University of Washington}}
     \city{Seattle}
      \state{WA}
      \country{US}
    }\email{yunhe@uw.edu}
    \author{Yongfeng Zhang}
    \affiliation{%
     \institution{\small{Rutgers University}}
     \city{New Brunswick}
      \state{NJ}
      \country{US}
    }\email{yongfeng.zhang@rutgers.edu}
    \author{Chirag Shah}
    \affiliation{%
     \institution{\small{University of Washington}}
     \city{Seattle}
      \state{WA}
      \country{US}
    }\email{chirags@uw.edu}

\begin{abstract}
As recommender systems become increasingly sophisticated and complex, they often suffer from lack of fairness and transparency.
Providing robust and unbiased explanations for recommendations has been drawing more and more attention as it can help address these issues and improve trustworthiness and informativeness of recommender systems.
However, despite the fact that such explanations are generated for humans who respond more strongly to messages with appropriate emotions, there is a lack of consideration for emotions when generating explanations for recommendations.
Current explanation generation models are found to exaggerate certain emotions without accurately capturing the underlying tone or the meaning.
In this paper, we propose a novel method based on a multi-head transformer, called Emotion-aware Transformer for Explainable Recommendation (EmoTER), to generate more robust, fair, and emotion-enhanced explanations.
To measure the linguistic quality and emotion fairness of the generated explanations, we adopt both automatic text metrics and human perceptions for evaluation.
Experiments on three widely-used benchmark datasets with multiple evaluation metrics demonstrate that EmoTER consistently outperforms the existing state-of-the-art explanation generation models in terms of text quality, explainability, and consideration for fairness to emotion distribution.
% Crowd-sourced experiments show that EmoTER increases the robustness of emotion-aware explanation without losing much of the informativeness, friendliness, credibility, quality, human-likeness, and even behaves better in certain dimensions. 
Implementation of EmoTER
will be released as an open-source toolkit to support further research.

% original Abstract
% As recommender systems become increasingly sophisticated and complex, they are more likely to suffer from the lack of fairness and transparency.
% Providing human-like explanations for recommendations can improve trustworthiness and informativeness in the system.
% However, emotion, one of the most critical factors in natural human communication, is still under-explored when generating explanations for recommendations.
% In this paper, we argue that explanations incorporating emotional content and factual information are perceived as more humane and more valuable.
% We propose a novel method based on multi-head transformer, called Emotion-aware Transformer for Explainable Recommendation (EmoTER), to generate emotion-aware explanations.
% Experiments on three different datasets with multiple metrics demonstrate that EmoTER outperforms the baselines in terms of emotion consistency, text quality, and explainability.
% In addition, we conduct crowd-sourced experiments to evaluate the usefulness of explanations generated by EmoTER.
% Our results show that the emotion-aware explanations are better perceived by users for their credibility and quality.
% The emotion-aware explanation generation model proposed and validated in this work opens up a new direction for AI systems with interactive elements through natural language interfaces.
\end{abstract}

\iffalse

\fi

%% A ``teaser'' image appears between the author and affiliation
\maketitle
\pagestyle{plain}

\section{Introduction}
Recommender systems assist users in finding relevant, contextual information and making decisions in diverse domains, such as shopping, entertainment, education, and healthcare.
Advances of deep learning and big data~\cite{zhang2014explicit} have made recommender systems increasingly complex and opaque, which hurts the trustworthiness and confidence in these systems~\cite{Zhang2018a}.
Incorporating explainability~\cite{herlocker2000explaining} has been explored as one of the reasonable ways to mitigate such concerns and increase transparency in decision-making processes.
% Providing post-hoc explanations for recommendation draws increasing attention in recent research studies.
Natural language sentence explanation, as a type of explanation that is easily understandable by average users, has gained much attention in recent years.
In particular, many text generation methods such as LSTM~\cite{6795963} and Transformers~\cite{Vaswani} have been applied to explanation generation models.
The Neural-Template (NETE) generation method~\cite{Li2020} integrates both template and generation-based approaches to make the explanation generation process more controllable.
PETER~\cite{li2021personalized} proposed a Personalized Transformer to generate personalized explanation sentences.
These works mainly focus on the text quality~\cite{papineni-etal-2002-bleu,lin-2004-rouge} and explainability ~\cite{Li2020,li2021personalized}. 

\begin{figure}[t]
 \centering
 \includegraphics[width=0.68\linewidth]{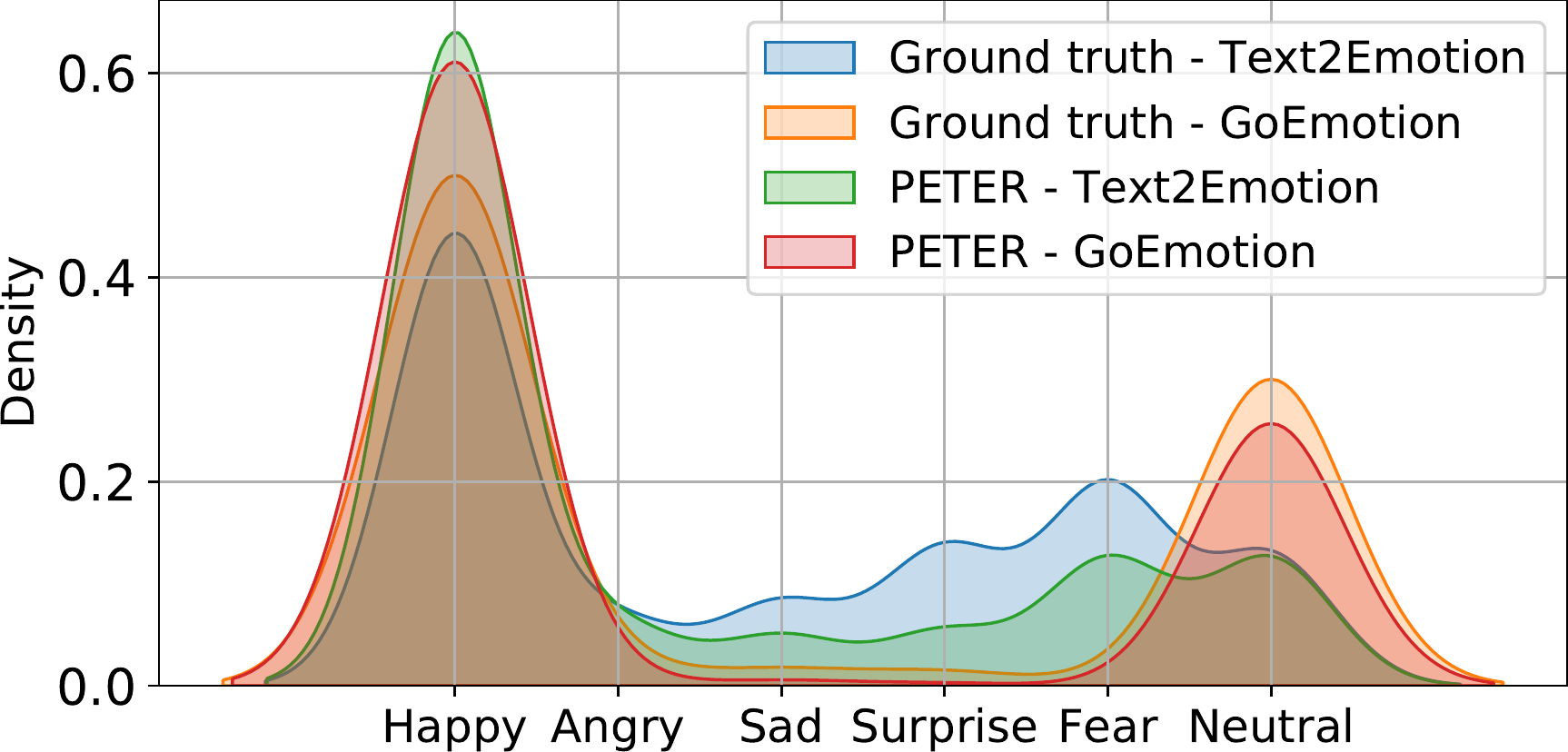}
 \vspace{-10pt}
 \caption{Existing state-of-the-art models generate biased emotion-aware explanations. For example, PETER~\cite{li2021personalized} generated more ``happy'' explanations but less ``sad'' and ``fear'' explanations than the ground truth based on two emotion detectors -- Text2Emotion~\cite{Ni} and GoEmotion~\cite{Demszky}.}
 \label{fig:distribution_emotion_bias}
 \vspace{-15pt}
\end{figure}

High-quality explanations should be informative, i.e., containing factual information for users to understand. In addition, the generated explanations should also sound natural and be comprehensible as well as empathetic towards users. Monotonous and dull sentences could make users feel disengaged and less trustworthy~\cite{Huang}. 
Emotion-aware text could address these issues by diversifying explanations and making them more useful and trustworthy to the users. 
Effects of emotion-induced messages to comprehension and trust has been well established in fields such as psychology, philosophy, and cognitive science (e.g., \cite{rorty1978explaining, ford1992motivating, parkinson1996emotions}. Areas where emotion-induced content has been applied include text generation ~\cite{Singh,10.1007/978-3-642-24571-8_2}, conversations \cite{fiehler2002emotions}, and social media \cite{kim2012you}. However, thus far web search or recommendation systems scholars have not focused on emotional aspects of explanations and their impacts on informativeness or usefulness in providing explainability for recommendations. This is a novel problem as it is challenging, because we need to pay attention to or extract emotional information while retaining the original meaning of the explanation.

To explore the emotional aspect of explanations, we selected six widely-used emotion categories: happy, angry, surprise, sad, fear, and neutral in recommendation scenarios inspired by emotion theories proposed by ~\citet{plutchik1962emotions,Plutchik1994ThePA}.
These are also at the intersection of common emotion categories used in the research work, such as emotion recognition~\cite{Demszky, zhou2018emotional, Huang} and emotion text generation~\cite{Singh,10.1007/978-3-642-24571-8_2}. Table~\ref{tab:emotion_category} shows some explanation examples corresponding to different emotion categories. 

When examining datasets and generative models for recommendation explanations, we find certain emotions are over-represented and considerable emotion biases exist, leading to lack of fairness as per the disparate impact definition \cite{singh2018fairness}.
For instance, in the TripAdvisor dataset, one of the most commonly-used recommendation explanation datasets, the ``happy'' emotion dominates others significantly.
To be specific, 42.8\% and 59.2\% explanations in TripAdvisor are classified as ``happy'' by emotion detector Text2Emotion~\cite{Ni} and GoEmotion~\cite{Demszky}, respectively (see the ground-truth distributions in Figure~\ref{fig:distribution_emotion_bias}).
More importantly, we also observe that state-of-the-art models (e.g., PETER~\cite{li2021personalized}) amplify such emotion bias in generated explanations and fail to perform robustly regarding different emotions.
As shown in Figure~\ref{fig:distribution_emotion_bias}, 61.3\% and 69.5\% explanations generated by PETER are classified as ``happy'' by Text2Emotion and GoEmotion respectively, leading to less ``sad'', ``surprised'', ``fear'' explanations than the ground-truth TripAdvisor dataset, which implies that the explanation generator could be unproportionally generating pleasant explanations to attract or even cheat users into the recommendations, and this hinders the trustworthiness of the explanations and the recommender system.

\begin{table}[t]
\centering
 \caption{Explanation examples on six emotion categories on the TripAdvisor dataset.}
 \label{tab:emotion_category}
 \vspace{-10pt}
 \begin{tabular}{c|p{0.8\columnwidth}}
 \toprule
 \toprule
 Emotion & Explanation\\
 \midrule
 Happy & on the rooftop an open sky swimming pool - fun \\
 \midrule
 Angry & Shame on you for your terrible customer service  \\
 \midrule
 Surprise & The big surprise is the fantastic dining room for lunch with stations for every food  \\
 \midrule
 Sad & The restaurant staff was nice but the buffet breakfast was pretty sad for the 18\\
 \midrule
 Fear & It was a miserable way to end a trip to a great country \\
  \midrule
 Neutral & the hotel is directly opposite budapest famous chain bridge and the danube \\ 
 \bottomrule
\end{tabular}
\vspace{-15pt}
\end{table}

In order to address the robustness and emotion bias problem in explanation generation, in this paper, we take the emotion factor into account and propose a novel method based on a multi-head transformer, called Emotion-aware Transformer for Explainable Recommendation (EmoTER). EmoTER can use explicit contextual information on emotions to generate emotion-aware explanations with high text quality, explainability, and robustness. Specifically, we extract the contextual emotion information as a guiding signal and introduce an emotion-aware module to explicitly integrate the emotion information during the encoding process.
% prevent emotion from vanishing during the encoding process. 
Then we add an emotion recognition head in the decoding process that enables emotion constraints.
To measure the linguistic quality and emotion fairness of the generated explanations, we adopt automatic text metrics and human perceptions for evaluation. 

Our main contributions in this paper are:
\begin{itemize}[leftmargin=*]
\item{[Conceptual] We first address the emotion bias issues on three benchmark datasets and current state-of-the-art models for explainable recommendation and propose an emotion-aware generation framework to generate robust and fair explanations. To the best of our knowledge, this is the first work to explore robust, fair, emotion-aware explanation generation in explainable recommendations.}
 \item{[Theoretical] We propose EmoTER (Emotion-aware Transformer for Explainable Recommendation), a novel multi-head Transformer architecture, which enhances the emotion of the explanations with a multi-task learning approach for personalized explanation generation.}
 \item{[Empirical] Experiments on three widely-used benchmark datasets with multiple evaluation metrics demonstrate that EmoTER consistently and significantly outperforms existing state-of-the-art explanation generation models in terms of robustness of emotion distribution, text quality, and explainability.
%  Crowd-sourced experiments show that the EmoTER increases the robustness of emotion-aware explanation without losing much of the informativeness, friendliness, credibility, quality, human-likeness, and even behaves better in certain dimensions.
 We also demonstrate that the proposed method leads to more equitable distribution of emotions in explainable, providing enhanced fairness based on disparate impact definition \cite{singh2018fairness}.}
\end{itemize}

\section{Related Work}
We begin by summarizing the related literature about explainable recommendation, emotional text generation and robustness of natural language generation in this section.

\noindent\textbf{Explainable recommendation}. Explainable recommendation has two major research perspectives: machine learning (ML)~\cite{zhang2014explicit,Zhang2018a,chen2021generate,chen2016learning} and human-computer interaction (HCI)~\cite{Google,ren2017social}. 
Our work incorporates both of these perspectives with a new model to generate emotion-aware explanations as well as a set of methods for evaluating such explanations using various metrics, including human judgments.  
%Our work makes contributions to both ML and HCI perspectives by proposing new algorithms for generating emotion-aware explanations and performing a user study to evaluate the explanations. 
%The former, provides explanations by designing new explainable recommendation algorithms, while the latter investigates how people perceive various styles of explanations. 
From the ML perspective, the methods for providing natural language explanations, which are most related to this work, can be categorized into template-based and generation-based methods. Pre-defined templates~\cite{zhang2014explicit} are not only simple and easy to use but also intuitive. However, designing templates is labor intensive and may limit the flexibility of the explanations. Recently, we have seen techniques involving LSTM~\cite{6795963} and Transformers~\cite{Vaswani} making great strides in natural language explanation generation. For instance, Chen et al.~\cite{Chen_Zhang_Qin_2019} proposed time-aware GRU to model dynamic user preferences. The Neural-Template (NETE) method~\cite{Li2020} integrates both template and generation-based approaches to make the explanation generation process more controllable. PETER ~\cite{li2021personalized} is a Personalized Transformer to generate personalized explanation sentences. A big shortcoming of these works is that they primarily focused on the text quality~\cite{papineni-etal-2002-bleu,lin-2004-rouge} and explainability ~\cite{Li2020,li2021personalized}, leaving the emotional aspects of explanations under-explored, which motivates our work.

%Since explanations have various forms, researchers adopt many approaches to evaluate them. 
Generating and presenting explanations is not sufficient; we also need to assess if these explanations are effective. Explanations should serve to improve the transparency, persuasiveness, effectiveness, trustworthiness, efficiency, scrutability and user satisfaction \cite{Zhang2018a}. There are several methods and metrics for assessing generated explanations. 
An ideal way of evaluating the explainability of machine generated explanations is through an online user study. For instance, Balog et al~\cite{Google} measured recommendation explanation quality by collecting users' judgments on seven pre-designed goals. But doing such evaluations can be expensive, time-consuming, and challenging at scale. Therefore, offline evaluation is often a more suitable solution for general research scenarios. The most commonly used metrics for evaluating machine generated explanation sentences are BLEU~\cite{papineni-etal-2002-bleu}, METEOR~\cite{banerjee2005meteor} and ROUGE~\cite{lin-2004-rouge}. In this paper, we adopt both automatic metrics and human evaluation to access the emotion-aware explanations.

\noindent\textbf{Emotional Text Generation}. Although research works on generating emotion-aware explanations are scant, there are several advancements made in emotion-inclusive natural language generation. For instance, \citet{10.1007/978-3-642-24571-8_2} used syntactic rules to generate emotional sentences. Since the corpus used for training was relatively small, this technique was only able to generate simple emotional sentences.
\citet{Colombo2019AffectDrivenDG} used a sequence-to-sequence approach and emotion vector representation, and proposed a penalty for generating words with different emotions. \citet{Ghosh} came up with Affect-LM, which is a LSTM-based model and able to generate sentences in four emotion categories including positive, anxious, sadness, and anger. 

% and fails to generate grammatical text. 
Another area where emotional text generation has been very important is conversational agents as several recent works have investigated incorporating emotions into conversational agents (e.g., ~\cite{asghar2017affective}). Emotional Chatting Machine~\cite{zhou2018emotional} takes sequence-to-sequence framework to create emotion-rich information in the context of dialogue scenarios. It utilizes eight emotion categories such as anger, disgust, fear, happiness, like, sadness, surprise, and unknown. Other researchers \cite{Huang} have adopted transfer learning to deal with emotional generation problems. They use pre-trained language models such as BERT~\cite{devlin2019bert} and GPT-2~\cite{radford2019language}, which have achieved reasonably high performance. However, these models are typically very large and take a long time to train.

%Zhou et al., 2018; Huang et al., 2018; Ghosh et al., 2017
%There has been a lot of research in this direction but the problem of integrating state-of-the-art explanation generation models with affective information remains an area ripe for exploration. 
In contrast, our proposed model is very light-weight,
% builds on state-of-the-art approaches for neural language modeling, 
utilizes no prior syntactic knowledge and no pre-trained model, but is able to generate expressive emotional explanations.

\noindent\textbf{Robustness in Natural language Generation} Robustness failures sometimes result from dataset biases introduced during data collection or human labeling, affecting model generalization and model performance. For example, Lewis et al.~\cite{lewis-etal-2021-question} showed that if there is a considerable overlapping on the test dataset in open-domain question answering, many QA models will not get memorized from training data and exhibit much less accuracy. In natural language inference, McCoy et al.~\cite{mccoy-etal-2019-right} demonstrated that crowd-sourced datasets used for training natural language inference models might introduce specific patterns more easily detected by statistical learning. Moreover, Bras et al.~\cite{pmlr-v119-bras20a} proposed light-weight adversarial filtering to filter dataset bias.
% Connection with Dataset Biases

Apart from finding the reasons for robustness failures, we should increase the robustness of the models with better use of minority examples. For example, Yaghoobzadeh et al.~\cite{yaghoobzadeh-etal-2021-increasing} proposed fine-tuning the model on full data and then on minority data. DRO (Distributional Robust Optimization)~\cite{Sagawa*2020Distributionally} addressed the training strategy on particularly hard examples. There are lots of extensions for DRO, e.g., Nam et al.~\cite{nam2020learning} emphasized the model's early-stage decision-making behaviors to train another model. Lahoti et al.~\cite{lahoti2020fairness} adopted additional model to recognize examples. Liu et al.~\cite{pmlr-v139-liu21f} proposed to weight minority examples that have high training loss.

Inspired by previous work on robustness, we propose an emotion-aware module to better use minority examples such as training examples belonging to some under-represented emotion categories. This will also enhance disparate impact view of fairness \cite{singh2018fairness}, which aims to provide representation of data proportional to the underlying distribution of those categories.
% Training with a better use of minority examples 
% https://arxiv.org/pdf/2112.08313.pdf

%As robustness gained increasing attention in NLP literature, various lines of work have proposed ways to identify robustness failures in NLP models. Existing works can be roughly categorized by how the failures are identified, among which a large portion of work relies on human priors and error analyses over existing NLP models, and a few other lines of work adopt model-based approaches. The identified robustness failure patterns are usually organized into challenging/adversarial benchmark datasets to more accurately measure an NLP model's robustness. 

% \vspace{-1cm}
\section{Methodology}
In this section, we formulate the problem of the emotion-aware explanation generation, describe the input representation, and present the detailed design and implementation of EmoTER.
% \vspace{-0.3cm}
\subsection{Problem Formulation}
We aim to propose a model $\mathcal{M}$ to generate emotion-aware explanations for recommendations.
We denote the generated recommendation explanation for the user $u$ and the item $i$ as $EXP_{u,i} = e_1,..., e_{E_{u,i}}$, where $e_1,...e_{E_{u,i}}$ are the explanation's word sequence and ${E_{u,i}}$ denotes the number of explanation words.
Item features in the ground-truth datasets are represented by $T_{u,i}$.
We take the user $u$ and item $i$ as the input $S=[u,i,f_{T_1},...f_{T_{F_{u,i}}},f_{emo},e_1,...e_{E_{u,i}}]$, where $f_{T_1},...f_{T_{F_{u,i}}}$ are topic features and $f_{emo}$ is emotion tag feature. ${F_{u,i}}$ represents the number of topic features. The output of context encoder would be $hidden_{context}$.

% We aim to propose a model $\mathcal{M}$ to generate emotion-aware explanations for recommendations.
% we define the generated recommendation explanation as $EXP_{u,i}$.
% Item features in the ground truth datasets are represented by $T_{u,i}$.
% We take the user $u$ and item $i$ as the input $S=[u,i,f_{T_1},...f_{T_{F_{u,i}}},f_{emo},e_1,...e_{E_{u,i}}]$, where $f_{T_1},...f_{T_{F_{u,i}}}$ are topic features and $f_{emo}$ is emotion tag feature. $e_1,...e_{E_{u,i}}$ are the explanation’s word sequence.${F_{u,i}}$ represents the number of topic features and ${E_{u,i}}$ denotes the number of explanation words.
% The output of context encoder would be $hidden_{context}$.

\begin{figure}[t]
 \centering
 \begin{minipage}[bt]{0.5\textwidth}
  \centering
 \includegraphics[width=\linewidth]{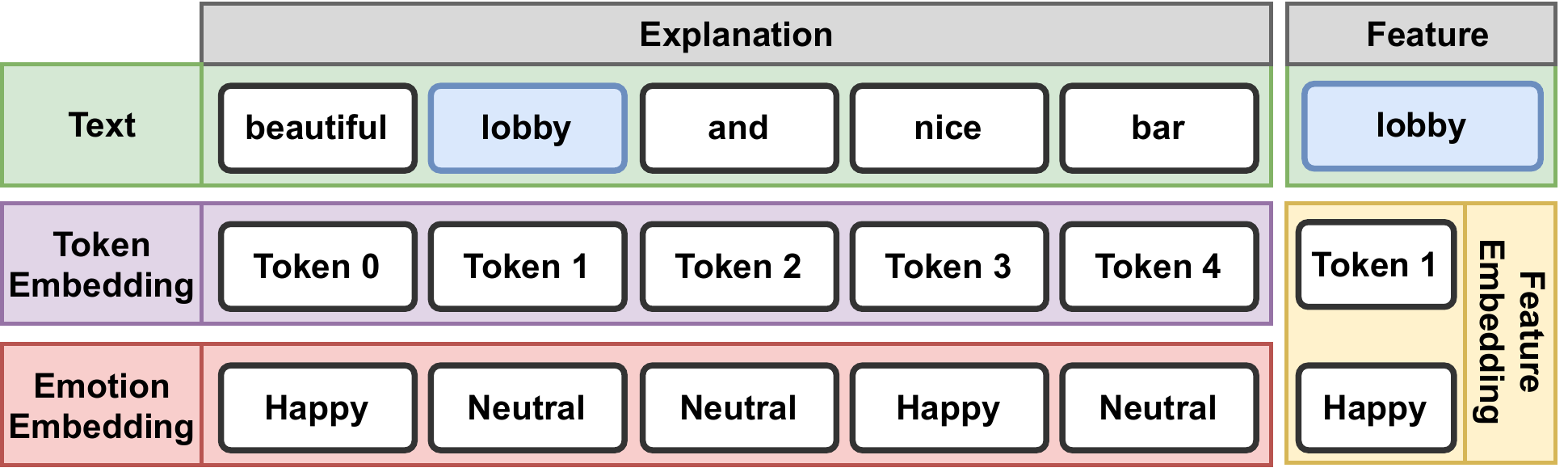}
 \vspace{-10pt}
 \caption{Input representation of the explanation \textit{``beautiful lobby and nice bar''} generated for the item \textit{hotel}, with \textit{lobby} serving as the feature of item \textit{hotel}.} 
 \label{fig:input_rep}
 \end{minipage}
% \hspace{0.3cm}
%  \begin{minipage}[bt]{0.45\textwidth}
% \centering
%  \caption{Statistics of the three datasets. *exp means explanation.}
%  \label{tab:dataset}
%  \begin{tabular}{ccccccl}
%  \toprule
%  \toprule
%  & TripAdvisor& Yelp & Amazon & \\
%  \midrule
%  \#user&9765 &27147& 7506\\
%  \#item&6280&20266& 7360 \\
%  \#features&5069&7340& 5399 \\
%  \#records&320023&1293247& 441783 \\
%  \#records/user&32.77&47.64&58.86 \\
%  \#records/item&50.96&63.81&60.02 \\
%  \#word/exp&13.01&12.32&14.14 \\
%  \bottomrule
% \end{tabular}
%  \end{minipage}
\vspace{-10pt}
\end{figure}

\begin{figure*}[t]
 \centering
 \includegraphics[width=\linewidth]{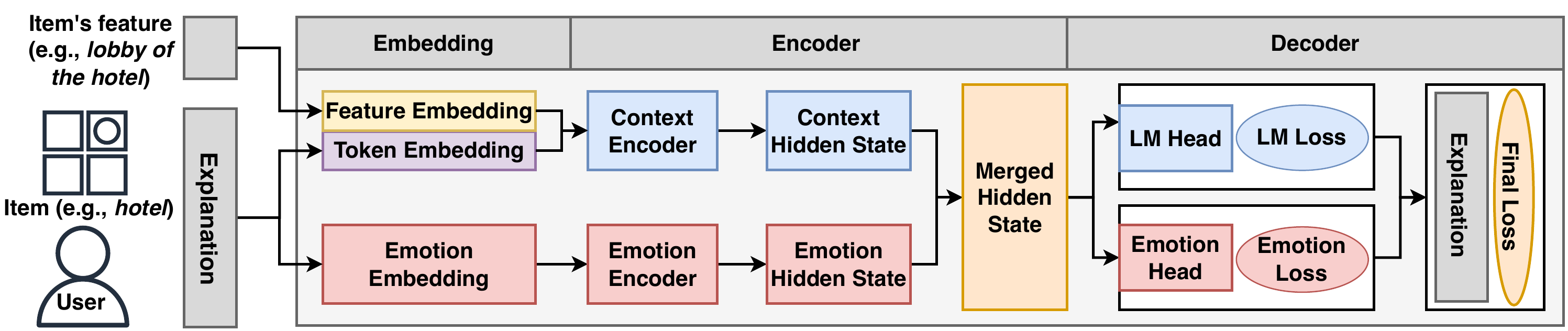}
 \vspace{-15pt}
 \caption{The EmoTER architecture overview. EmoTER takes three types of embeddings (i.e., feature embedding, token embedding, and emotion embedding) as inputs and encode them into context and emotion hidden states. Then we merge the two hidden states and apply a two-head (emotion head and language modeling head) decoder to generate explanations.}
 \label{fig:framework}
 \vspace{-5pt}
\end{figure*}

% The objective of our explanation task is to control the output $E_{u,i}$ of the explanation generation model with emotional information for a pair of user
% $u$ and item $i$. Item features provided by the ground truth is considered as topic $T_{u,i}$ of the expected explanation, and the rating $R_{u,i}$ is also helpful by showing the intensity of the expressed emotion.
% \vspace{-0.3cm}
\subsection{Input Representation}
As shown in Figure~\ref{fig:input_rep}, the input representation consists of three parts: emotion embedding, token embedding, and feature embedding.
All the three embeddings are fed into the transformer encoder layers, which encode the context information of explanations with an embedding size of 512.
We use positional embedding with a length of 512, which is the same as the embedding size. 

% Figure~\ref{fig:inputemb} demonstrates the input representation. The embeddings consist of three parts: token embedding, emotion Token embedding and feature embedding. All the encoder layers use positional embeddings with a length of 512, which is the same as embedding size.

\subsubsection{Emotion Embedding} We construct the word emotion embedding based on the emotion category and intensity extracted from NRC Emotion Lexicon~\cite{LREC18-AIL}, which is a word dictionary containing words and their corresponding emotion.
When creating the word emotion embedding, we consider six most commonly used emotions including anger, fear, trust, surprise, sad, and happy.
% such as anger, fear, anticipation, trust, surprise, sadness, joy, and disgust. When creating the word emotion embedding, we first select six out of the above eight basic emotion by removing ``anticipation'' and ``disgust'' since we adopt the most commonly-used emotion categories. 
For example, the representation of word ``lucky'' is expressed as a 6-dimension vector $V_{NRC}$, which equals to \{happy:0.721, angry:0, surprise:0.539, sad:0, fear:0, neutral:0\}.
If we cannot find a word in the NRC Emotion Lexicon, then we assign a vector \{happy:0, angry:0, surprise:0, sad:0, fear:0, neutral:1.0\} to it.

% \begin{itemize}
%  \item Emotion embedding: We obtain the emotion category and intensity from NRC Emotion Lexicon~\cite{LREC18-AIL}. It is a list of English words and their associations with eight basic emotions (anger, fear, anticipation, trust, surprise, sadness, joy, and disgust) and two sentiments (negative and positive). The annotations were manually done by crowdsourcing. For example, the representation of word `'lucky'' is 6-dimension vector $V_{emo}$, which equals to \{happy:0.721, angry:0, surprise:0.539, sad:0, fear:0, neutral:0\}.  If we cannot find a word in the NRC Emotion Lexicon, then we will assign vector \{happy:0, angry:0, surprise:0, sad:0, fear:0, neutral:1.0\} to it.

Once we have constructed a 6-dimension vector $V_{NRC}$, we convert it into a 512-dimension emotion embedding $V_{emo}$ using Eq.~\ref{equ:perceptron}.
\begin{equation}~\label{equ:perceptron}
   %V_{emo} =  (R_{ui}-R_{avg})*V_{preemo}*W^{6*512}
    V_{emo} =  g(V_{NRC})
\end{equation}
where $g(.)$ is a multi-layer perceptron with an input of 6 and an output size of 512.

%  \begin{equation}
%   %V_{emo} =  (R_{ui}-R_{avg})*V_{preemo}*W^{6*512}
%  V_{emo} =  g(V_{preemo})
% \end{equation}
% where $V_{emo}$ is 6-dimension pre-trained embedding obtained from NRC emotion lexicon and g(.) is the output of a multi-layer perceptron operating on $V_{pre_emo}$.

\subsubsection{Token Embedding} We use a simple tokenization method and treat a word as a token.
The vocabulary used in our study consists of 20,000 most frequent tokens.
There are 10 special tokens including four functional words: $\langle$bos$\rangle$ (begin of sentence), $\langle$eos$\rangle$ (end of sentence), $\langle$pad$\rangle$ (padding word), $\langle$unk$\rangle$ (unknown word) as well as six frequently used emotion category words to denote word emotions: $\langle$happy$\rangle$, $\langle$angry$\rangle$, $\langle$surprise$\rangle$, $\langle$sad$\rangle$, $\langle$fear$\rangle$, $\langle$neutral$\rangle$. 
% For example, $\langle$bos$\rangle$ indicates the beginning position of the explanation, and $\langle$eos$\rangle$ denotes the end position of the explanation.

\subsubsection{Feature Embedding} We integrate item features and emotion tags to construct feature embeddings, as shown in the orange box of Figure~\ref{fig:input_rep}.
The item feature ${f_{item}}$ is self-contained in the dataset, while the emotion tag ${f_{emo}}$ is assigned by an external emotion classification detector.
% based on ground-truth explanation sentences.
In our study, we choose a widely-used emotion classifier named Text2Emotion~\cite{koper-etal-2017-ims} to create emotion tags.

% \vspace{-0.3cm}
\subsection{Model Description}
EmoTER adopts two encoders and two-head decoder based Transformer architecture and takes as input a sequence in the embedding space of emotion embeddings, token embeddings, and feature embeddings, as shown in Figure~\ref{fig:framework}.
To be specific, one encoder takes emotion embeddings as input and creates the emotion hidden state, while the other encoder is responsible to process the concatenated feature embeddings and token embeddings to generate the context hidden state.
Next, the two hidden states are fused by vector summation.
We design a two-head decoder, i.e., the emotion head and the language modeling head, to project the last layer of the merged hidden state to two emotion and language modeling representations with desired dimensions.
Accordingly, the final loss is expressed as a weighted sum of the two losses based on the emotion and language modeling constraints. 

% We design two decoder heads, i.e, the emotion head and the language modeling head, to project the last layer of the merged hidden state to two vectors with desired dimensions: a 6-dimension emotion vector and a 15-dimension language vector.
% The loss functions includes language modeling and emotion constraints, as shown in Figure~\ref{fig:framework}.
% The final loss is a weighted sum of the two losses.

EmoTER is designed based on the framework of PETER~\cite{li2021personalized} by adding one more encoder and two feed-forward linear heads on decoder .
For the two encoders in EmoTER, we adopt a 2-layer architecture because: (1) it is fair to compare with PETER's single 2-layer encoder \cite{li2021personalized}; (2) a 2-layer encoder is light-weight and easy to apply into different tasks.
Each Transformer encoder layer is composed of a multi-headed self-attention and a position-wise feed-forward network.
In the following, we introduce these two encoders and two different heads and their corresponding loss functions.

% EmoTER is a two-head encoder and two-head decoder Transformer architecture. We will now describe the full pipeline of EmoTER. Our model takes as input a sequence in the embedding space including token embedding, emotion embedding and feature embedding. Two head encoder takes feature and token, emotion respectively. Then we obtain the hidden state from both encoders and pass the two hidden states into a function $F$. We call the hidden state that passing through $F$ function, $F$ hidden state. We add two decoder heads to the Transformer, which projects the last layer of $F$ hidden state to the output size. The loss functions includes language modeling and emotion constraints, as shown in Figure~\ref{fig:framework}. The final loss is a weighted sum of all the losses. EmoTER is built upon the framework of PETER~\cite{li2021personalized}. We add one encoder and two feed-forward linear heads on top of the PETER losses.
% The two head encoder module use 2-layer architecture, which is the same as PETER for better comparison. To be light-weight and easy to apply into different specific task, we use a 2-layer architecture. In the following, we define these two different heads and their corresponding loss functions.

%%%%%%%%%%%%%%%%%%%%%%%%%%%%%%%%%%%%%%

\subsubsection{Emotion Encoder} The emotion encoder serves as an emotion-aware module in EmoTER to encode the emotion information when processing the input.
% the prevention of emotion information vanishing.
% As Figure~\ref{comparison} shows, 
In particular, the emotion-aware module guides EmoTER to pay more attention to the emotional words and strengthen the weights on them, which explicitly encodes the emotional information and enables emotion-guided decoding for explanation generation.
% mitigating emotional information vanishing during the decoding process.

The input sequence of the user $u$ and the item $i$ to emotion transformer can be expressed as:
\begin{equation}
S=[u,i,emo_{T_1},...,emo_{T_{F_{u,i}}},emo_{f_{emo}},emo_{e_1},...,emo_{e_{E_{u,i}}}]
\end{equation}
where $emo_{T_i}$ means the emotion embedding of $i_{th}$ item feature. ${F_{u,i}}$ represents the number of item features and ${E_{u,i}}$ denotes the number of explanation words.
We denote the hidden state of emotion encoder as $hidden_{emo}$.

% \subsubsection{Emotion encoder} We also call it emotion aware module. This module is vital to our task, which prevents emotion information vanishing in the context encoder. As Figure~\ref{comparison} shows, emotion aware module provides guidance to attend emotional words and strengthen the weights on them, which prevent emotional information vanishing during the decoding process. Input sequence to emotion transformer can be represented as 
% \begin{equation}
% S=[u,i,emo_{T_1},...,emo_{T_{F_{u,i}}},emo_{tag},emo_{e_1},...,emo_{e_{E_{u,i}}}]
% \end{equation}
% where all words change their embedding method from token embedding to emotion embedding.
% The output of emotion encoder would be $hidden_{emo}$.

% Fusion function is computed as:
% \begin{equation}
% hidden_merge = intensity * hidden_{emo} + hidden_{context}
% \end{equation}
% Here, $hidden_{merge}$ denotes as context representation of our model. Intensity works as modifier to adjust the strength of emotion. 

%%%%%%%%%%%%%%%%%%%%%%%%%%%%%%%%%%%%%%
\subsubsection{Context Encoder}~\label{sec:Context_encoder}
Input sequence to context transformer can be expressed as:
\begin{equation}
S=[u,i,f_{T_1},...f_{T_{F_{u,i}}},f_{emo},e_1,...e_{E_{u,i}}]
\end{equation}
where $f_{T_1},...f_{T_{F_{u,i}}}$ are topic features and $f_{emo}$ is emotion tag feature. $e_1,...e_{E_{u,i}}$ are the explanation's word sequence. ${F_{u,i}}$ represents the number of item features and ${E_{u,i}}$ denotes the number of explanation words.
The output of context encoder would be $hidden_{context}$.

When the emotion hidden state $hidden_{emo}$ and the context hidden state $hidden_{context}$ are ready, we merge them into $hidden_{merge}$ by:
\begin{equation}
hidden_{merge} = intensity * hidden_{emo} + hidden_{context}
\end{equation}
where the coefficient $intensity$ works as a modifier to adjust the strength of emotion. 

% \subsubsection{Context encoder}~\label{sec:Context_encoder} Each Transformer encoder layer is composed of multi-headed self-attention and position-wise feed-forward network.
% Input sequence to context transformer can be expressed as
% \begin{equation}
% S=[u,i,f_{T_1},...f_{T_{F_{u,i}}},f_{emo},e_1,...e_{E_{u,i}}]
% \end{equation}
% where $f_{T_1},...f_{T_{F_{u,i}}}$ are topic features and $f_{emo}$ is emotion tag feature. $e_1,...e_{E_{u,i}}$ are the explanation’s word sequence.${F_{u,i}}$ represents the number of topic features and ${E_{u,i}}$ denotes the number of explanation words.
% The output of context encoder would be $hidden_{context}$.

%%%%%%%%%%%%%%%%%%%%%%%%%%%%%%%%%%%%%%
\subsubsection{Emotion Head}
Emotion head is trained to perform the emotion-constrained classification task.
We add this head to constrain the model to focus on the emotion information expressed in the context.
The prediction probability of target emotion is computed as:
\begin{equation}
 P(emo_i|emo_1,emo_2,...,emo_t) = softmax(hidden_{merge}[-1]*M_1)
\end{equation}
where $emo_i$ is the predicted emotion category $i$, and $emo_1,...,emo_t$ are all emotion categories. 
$hidden_{merge}[-1]$ is the last hidden layer of our model and $M_1$ is the weight to be learned during the training for the emotion classification task.
Then we have the following cross-entropy loss:
\begin{equation}
 L_{emo}(E) = - \sum_{i=1}^{N}log(P(emo|emo_1,emo_2,...,emo_t))
\end{equation}

% \subsubsection{\note{Emotion \textbf{modeling} head}}. Emotion head is trained to perform emotion classification task. We add this head to constrain the model focus on the emotion information expressed in the context. The prediction probability of target emotion is computed as 

% \begin{equation}
%  P(emo_i|emo_1,emo_2,...,emo_t) = softmax(hidden_{merge}[-1]*M_2)
% \end{equation}

% where $emo_i$ is the predicted emotion category and $emo_1,emo_2,...,emo_t$ is all emotion categories.  $hidden_{merge}[-1]$ is the last hidden layer of the our model and $M_2$ is the weights to be learned during the training for the emotion classification task. Then we choose cross-entropy as our loss function:
% \begin{equation}
%  L_{emo}(E) = - \sum_{i=1}^{N}log(P(emo|emo_1,emo_2,...,emo_t))
% \end{equation}

%%%%%%%%%%%%%%%%%%%%%%%%%%%%%%%%%%%%%%
\subsubsection{Language Modeling Head} We add a 2-layer transformer decoder, which predicts the next token based on the merged context vector $hidden_{merge}$.
We define the explanation as $E = {e_1, e_2, ..., e_n}$, then the conditional probability of the next token is computed as:
\begin{equation}
 P(e_i|e_1,e_2,...,e_{i-1}) = softmax(hidden_{merge}[-1]*M_2)
\end{equation}
where $hidden_{merge}[-1]$ is the last hidden layer and $M_2$ is the token embedding matrix. During training, $M_2$ updates its weights. We also have the following cross-entropy loss:
\begin{equation}
 L_{lm}(E) = - \sum_{i=1}^{N}log(P(e_i|e_1,e_2,...,e_{i-1}))
\end{equation}
where $N$ is the word count in the generated explanation.

% \begin{figure}[t]
%     \subfigure[Attention map of PETER.
%       \label{fig:vis_peter}]{\includegraphics[width=0.35\linewidth]{figure/vis_peter.pdf}}
%       \hspace{2cm}
%     \subfigure[Attention map of our model.
%       \label{fig:vis_emoter}]{\includegraphics[width=0.35\linewidth]{figure/vis_emoter.pdf}}
%     \caption{Comparison of attention map from PETER and the proposed EmoTER when generating explanations from the same user item pair. Lighter the block is, higher attention score is. Our model EmoTER greatly pays more attention to emotional expressions.}
%     \label{comparison}
% \end{figure}

%%%%%%%%%%%%%%%%%%%%%%%%%%%%%%%%%%%%%%
\subsubsection{Multi-task Learning} Finally, we integrate the two tasks into a multi-task learning framework with the following loss function:
\begin{equation}\label{equ:loss_total}
L_{total} = c_1L_{lm} + c_2L_{emo}
\end{equation}
where $c_1$ and $c_2$ are two hyper-parameters to control the weights of emotion loss and language modeling loss.

% \begin{table}[t]
% \centering
%  \caption{Statistics of the three datasets. *exp means explanation.}
% %  \vspace{-0.3cm}
%  \label{tab:dataset}
%  \begin{tabular}{ccccccl}
%  \toprule
%  \toprule
%  & TripAdvisor& Yelp & Amazon & \\
%  \midrule
%  \#user&9765 &27147& 7506\\
%  \#item&6280&20266& 7360 \\
%  \#features&5069&7340& 5399 \\
%  \#records&320023&1293247& 441783 \\
%  \#records/user&32.77&47.64&58.86 \\
%  \#records/item&50.96&63.81&60.02 \\
%  \#word/exp&13.01&12.32&14.14 \\
%  \bottomrule
% % \vspace{-0.5cm}
% \end{tabular}
% \end{table}

\begin{table}[t]
\centering
 \caption{Statistics of the three datasets.}
%  *exp means explanation.}
 \label{tab:dataset}
 \vspace{-10pt}
 \begin{tabular}{ccccccl}
 \toprule
 \toprule
 & TripAdvisor& Yelp & Amazon & \\
 \midrule
 \#user&9765 &27147& 7506\\
 \#item&6280&20266& 7360 \\
 \#features&5069&7340& 5399 \\
 \#records&320023&1293247& 441783 \\
 \#records/user&32.77&47.64&58.86 \\
 \#records/item&50.96&63.81&60.02 \\
 \#word/explanation&13.01&12.32&14.14 \\
 \bottomrule
\end{tabular}
\vspace{-10pt}
\end{table}

\begin{table*}[t]
  \centering
  \begin{minipage}[b]{0.45\textwidth}
    \centering
 \caption{Emotion distribution of explanations evaluated by Text2Emotion TripAdvisor dataset, PETER and our model EmoTER.}  
 \vspace{-10pt}
 \label{tab:Text2Emotion}
 \begin{tabular}{ccccccl}
 \toprule
 \toprule
 & Ground Truth & PETER & EmoTER & Debiasing  \\
%   & Dataset & PETER+ & Our model & Debiasing  \\
 \midrule
 happy& 42.8\% &   61.3\% &  47.2\% & 21.9\%\\
 angry& 5.9\%&  5.4\%  &  4.8\% & -10.2\% \\
 surprise& 12.7\%&  5.0\% &  9.8\%  & 37.8\% \\
 sad& 7.7\% & 4.6\% &  5.8\% &10.4\% \\
 fear& 18.7\%&  11.7\% & 17.4\% &30.5\% \\
  neutral& 12.2\%&  11.8\% & 14.9\%  &25.4\% \\
 \bottomrule
\end{tabular}
 \end{minipage}
\hfill
  \begin{minipage}[b]{0.45\textwidth}
    \centering
\caption{Emotion distribution of explanations evaluated by GoEmotion on TripAdvisor dataset, PETER and our model EmoTER.} 
 \label{tab:Goemotion}
 \vspace{-10pt}
 \begin{tabular}{ccccccl}
 \toprule
 \toprule
%  & Dataset & PETER+ & Our model & Debiasing\\
 & Ground Truth & PETER & EmoTER & Debiasing  \\
 \midrule
 happy& 59.2\% &   69.5\% &  63.6\% & 10.0\% \\
 angry& 1.5\%&  0.3\%  &  0.5\% & 13.3\% \\
 surprise& 1.6\%&  0.3\% &  0.4\% & 6.3\% \\
 sad& 1.8\% & 0.6\% &  1.3\% & 38.9\% \\
 fear& 0.3\%&  0.05\% &  0.06\% &3.3\%\\
  Neutral& 35.6\%&  29.2\% & 34.2\% & 14.0\%\\
 \bottomrule
\end{tabular}
 \end{minipage}
 \vspace{-10pt}
\end{table*}

\section{Experiment Setup}
In this section, we introduce the datasets and metrics for evaluation.
The baselines and model training are also presented.

\subsection{Datasets}

For experimentation, we adopted three publicly available explainable recommendation datasets~\cite{Li2020,li2021extra,Ni}: TripAdvisor (hotel), Yelp (restaurant) and Amazon (movies, TV). Details of the datasets are shown in Table~\ref{tab:dataset}.
Each dataset is randomly split into training, validation, and testing sets with a ratio of 8:1:1 for 5 times.
Each user and each item contains at least one record in the training set.
Each record is comprised of a user ID ($u$), an item ID ($i$), an explanation ($Exp_{u,i}$), and at least one item feature ($F_{item}$).
The explanations are extracted from user reviews~\cite{li2021extra}.

\subsection{Evaluation Metrics}
We evaluate our model's ability of generating explanations with expressive emotion, text quality, and efficacy of explainability on the test set. Overall, evaluations of the explanation performance could be divided into two parts: automatic metrics and human evaluations.

\subsubsection{Automatic Evaluations}

We use three various automatic evaluation methods to evaluate the emotion fairness, text quality, and explainability of explanations generated by our model EmoTER.

\noindent\textbf{Explanations with emotion fairness and robustness} As reported earlier, we found there is huge bias about the emotion distribution both on three benchmark datasets and explanations generated by state-of-the-art models. Here, we equate this {\em bias} to lack of fairness, where {\em fairness} is considered to be based on disparate impact \cite{singh2018fairness}. Our goal is to reduce this bias and increase fairness by having our explanations exhibit emotions that match the underlying distributions in ground truth. We use the same emotion classifiers to measure the emotion distribution of explanations generated by our model EmoTER for fair comparison. Text2Emotion~\cite{koper-etal-2017-ims} is a widely-used emotion detection API working on the emotional words of text. While GoEmotions~\cite{Demszky} is the largest manually labeled dataset on English Reddit comments with high quality of 28 emotion categories including Neutral. Its BERT-based model is one of the state-of-the-art emotion recognition models. In order to show the robustness and fairness effect of our emotion-aware module, we chose these two very different models as our emotion classifiers to identify the emotion distribution of the explanations from the ground-truth dataset, generated by PETER, and generated by our model, respectively.

\begin{figure}[t]
  \centering
  \begin{minipage}[b]{0.49\textwidth}
    % \vspace{-1.5cm}
    \centering
    \subfigure[Survey on desktops\label{fig:survey_robust}]{\includegraphics[width=0.624\linewidth]{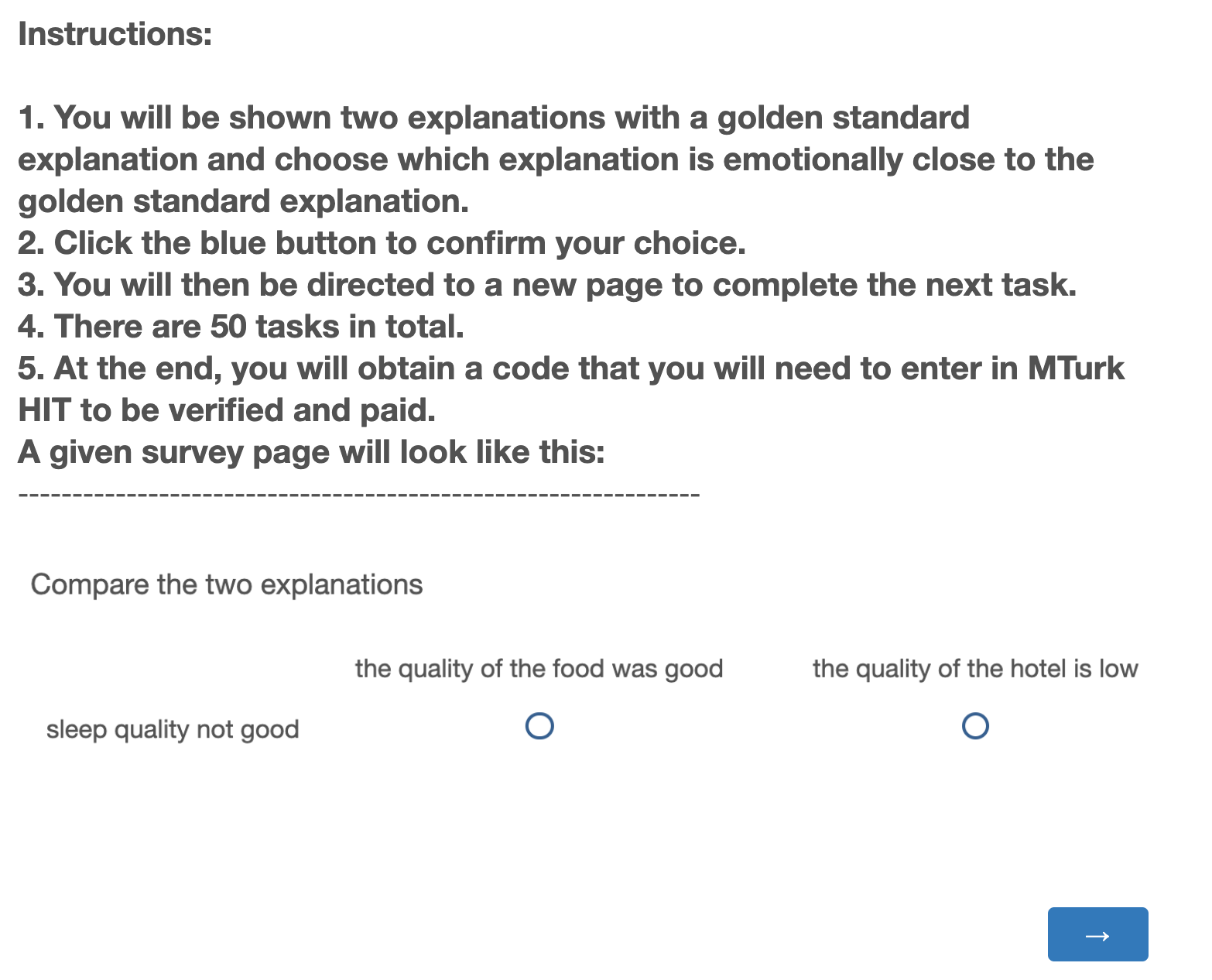}}
    % \hspace{1cm}
    \subfigure[Survey on smartphones\label{fig:survey_robust_mobile}]{\includegraphics[width=0.234\linewidth]{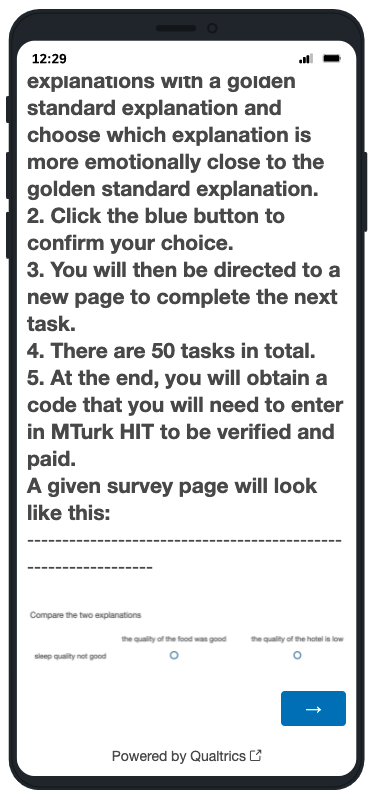}}
    % \vspace{-15pt}
    \caption{Survey on Amazon MTurk platform about evaluating explanations with emotion robustness and fairness. Explanations are generated by PETER and EmoTER respectively}
    \label{fig:user_survey}
  \end{minipage}
%   \hfill
%   \begin{minipage}[b]{0.49\textwidth}
%     \centering
%     \subfigure[Survey on desktops\label{fig:survey_desktop}]{\includegraphics[width=0.624\linewidth]{figure/survey_desktop.jpg}}
%     % \hspace{1cm}
%     \subfigure[Survey on smartphones\label{fig:survey_mobile}]{\includegraphics[width=0.234\linewidth]{figure/survey_mobile.jpg}}
%     \caption{Survey on Amazon MTurk platform about evaluating explanations on five dimensions: informativeness, credibility, friendliness, high quality, human-likeness.}
%     \label{fig:user_survey}
%   \end{minipage}
\vspace{-15pt}
\end{figure}

\noindent\textbf{Text quality} To assess the text quality of EmoTER-generated explanations and compare them with state-of-the-art models, we follow the widely-used automatic evaluation metrics. We adopt BLEU~\cite{papineni-etal-2002-bleu} and ROUGE~\cite{lin-2004-rouge}, which are commonly used in machine translation and in text summarization respectively. 
Since BLEU and ROUGE often fail to detect the problem of always generating simple and repetitive sentences such as ``I recommend it,'' 
% and causes explanations highly repetitive, 
we adopt USR (Unique Sentence Ratio)~\cite{Li2020} to measure the diversity of the generated explanations.
\begin{equation}
USR = Set(N)_{num}/N_{num}
\end{equation}
where $N$ is the collection of generated explanations and $Set(N)$ is the collection of unique explanations. We compare the similarity between explanations using exact matches.

\noindent\textbf{Explainability} To assess the explainability of EmoTER-generated explanations, we use metrics that are widely used in previous work. We adopt three metrics related to features in the datasets. Since people care about specific features of the recommended items~\cite{DBLP:journals/corr/abs-2101-03392}, we use three feature-based metrics: Feature Matching Ratio (FMR), Feature Coverage Ratio (FCR) and Feature Diversity (DIV), which are proposed in~\cite{Li2020}. 
\begin{equation}
FMR = N^f_{num}/G^f_{num}
\end{equation}
where $G^f$ is ground-truth explanation with feature and $N^f$ is generated explanations with feature. FMR measures generated explanations in terms of feature level.
\begin{equation}
FCR = F^N_{num}/F^G_{num}
\end{equation}
where $F^G$ is the distinct features in the ground-truth and $F^N$ is the distinct features in the generated explanations.

\begin{equation}
DIV = F_{u,i} \cap F_{u', i'}
\end{equation}
where $F_{u,i}$ and $F_{u', i'}$ are two sets of distinct features discussed in explanations of $u,i$ and $u', i'$ respectively. DIV measures the feature intersections between various $u,i$ pairs. Since different users may not always talk about the same feature about items, a lower DIV means better performance.

\subsubsection{Human Evaluations}
We conduct two different user studies to measure emotion fairness/robustness and quality of explanations from humans' perspectives. 

\noindent\textbf{Explanations with emotion fairness and robustness} We measure EmoTER's ability to generate fair and robust emotional explanations by conducting an extensive user study on Amazon's Mechanical Turk (MTurk) platform. As figure~\ref{fig:survey_robust} shows, we designed a survey to ask annotators to choose which explanation is more emotionally close to the ground-truth explanation. We presented a set of explanations generated using a baseline method PETER and the corresponding explanations generated using EmoTER in a randomly-ordered pair. The Turk workers did not know which explanation was emotion-aware. We used three sets of 50 such explanation pairs from three datasets each to create a HIT (Human Intelligence Task) on MTurk. Each HIT was done by three different Turk workers. Providing ratings on 50 items in this way took about 30 minutes. Each was paid 4.00 USD. This experiment was authorized by our Institutional Review Board (IRB) after reviewing it for ethical and privacy considerations. Each task was conducted by human raters that were located in the United States and had high approval rates.

% \noindent\textbf{Debiasing effect on general evaluation} In order to access the general quality of these two sets of explanations, we designed a survey to ask annotators to compare the quality of explanation-pair from five perspectives: informativeness, friendliness, credibility, quality, and human-likeness. We presented a set of explanations generated using a baseline method PETER+ and the corresponding explanations generated using EmoTER in a randomly-ordered pair. The Turk workers did not know which explanation was emotion-aware. The Turk workers were asked to pick which one of the given pairs they found higher on aspects of informativeness, friendliness, credibility, quality, and human-likeness. We used three sets of 50 such explanation pairs from three datasets each to create a HIT (Human Intelligence Task) on MTurk. Each HIT was done by three different Turk workers. Providing ratings on 50 items in this way took about 20 minutes. Each was paid USD 3. The same as the previous user study, this experiment was authorized by our Institutional Review Board (IRB) after reviewing it for ethical and privacy considerations. Each task was conducted by human raters that were located in the United States and had high approval rates. 

\begin{table*}[t]
\centering
%\vspace{-0.3cm}
 \caption{Performance comparison on three datasets TripAdvisor, Yelp and Amazon. R1 and R2 denote ROUGE-1 and ROUGE. P, R, F denote Precision, Recall and F1. The best-performing values are boldfaced. The second-performing values are underlined. BLEU and ROUGE are percentage values.}

 \label{tab:quantitative}
 \begin{tabular}{cc|ccc|*{9}{c}}
 \toprule
 \toprule
 & & \multicolumn{3}{c|}{Explainability} & \multicolumn{9}{c}{Text quality} \\
 \midrule
 Dataset & Model & FMR & FCR & DIV & USR & BLEU-1&BLEU-4&R1-P& R1-R&R1-F& R2-P& R2-R&R2-F \\
 \midrule
%  \midrule
%  & \multicolumn{12}{|c|}{TripAdvisor} \\
%  \midrule
 \multirow{4}{*}{TripAdvisor} & ACMLM &0.07&\underline{0.41}&\textbf{0.78}& \textbf{0.94} & 3.45 & 0.02 & 4.86 & 3.82 & 3.72& 0.18 & 0.20 &0.16 \\
 & NETE&0.79 & 0.27 & 2.23 & \underline{0.56} & 22.41&3.60&35.70& 24.68&27.70& 10.23& 6.98&6.55 \\
 & PETER &\textbf{0.89} & 0.36  & 1.59 & 0.25 & \underline{24.19}&\underline{4.54}&\underline{37.57}& \underline{29.15}&\underline{30.49}& \underline{11.99}& \underline{8.93}&\underline{9.27} \\
 & EmoTER & \underline{0.87} & \textbf{0.42} & \underline{1.52} & 0.39 & \textbf{25.14}& \textbf{4.89}&\textbf{39.29} & \textbf{30.04} & \textbf{31.76}& \textbf{13.12}& \textbf{9.64}&\textbf{10.08} \\
%   \midrule
%  & \multicolumn{12}{|c|}{Yelp} \\
 \midrule
  \multirow{4}{*}{Yelp} & ACMLM &0.05&0.31&\textbf{0.95}& \textbf{0.95}&7.01&0.24&7.89&7.54&6.82&0.44&0.48&0.39 \\
 & NETE&0.81 & 0.29 & 1.48 & \underline{0.52} & 19.34&2.63&33.95& 22.31&25.65& 8.88& 5.45&6.67 \\
 & PETER &\textbf{0.86} & \underline{0.36} & 1.07 & 0.30 & \underline{20.55} &\underline{3.40} & \underline{35.63} & \underline{25.93}& \underline{27.91} & \underline{10.76} & \underline{7.37}& \underline{7.96} \\ 
 & EmoTER &\textbf{0.86} & \textbf{0.38} & \underline{1.00} & 0.41 & \textbf{21.61} &\textbf{3.84}&\textbf{37.41}& \textbf{27.16}&\textbf{29.41}& \textbf{11.90}& \textbf{8.17}&\textbf{8.83} \\
  \midrule
%  & \multicolumn{12}{|c|}{Amazon} \\
%  \midrule
 \multirow{4}{*}{Amazon} & ACMLM &0.10 & \textbf{0.31} & 2.07 &\textbf{0.96} & 9.52 & 0.22 & 11.65 & 10.39 & 9.69 & 0.71 & 0.81 & 0.64 \\
 & NETE&0.69 & 0.19 & 1.95 & \underline{0.57} & 18.82& 2.47& 33.78& 21.31 &24.75& 7.64& 4.81&5.52\\
 & PETER &\textbf{0.77} & \underline{0.26} & \underline{1.25} & 0.39 & \underline{19.58}& \underline{3.00} & \underline{35.23} & \underline{23.84}& \underline{26.44} & \underline{8.99} & \underline{6.18}& \underline{6.63} \\ 
 & EmoTER & \underline{0.76} & \underline{0.26} & \textbf{1.22} & 0.46 & \textbf{20.20}&\textbf{3.21}&\textbf{35.56}& \textbf{24.47}&\textbf{27.06}& \textbf{9.32}& \textbf{6.57}& \textbf{7.03} \\
 \bottomrule
\end{tabular}
\end{table*}

% \subsection{Compared Methods}
\subsection{Baselines}
We considered three state-of-the-art explanation generation models as baselines in our experiments. 
\begin{itemize}
    \item NETE~\cite{Li2020} integrates template-based and generation-based approaches to make the explanation generation process more controllable.
    \item ACMLM \cite{Ni} is a BERT-based explanation generation model. ACMLM fine-tunes pre-trained BERT to encode both item and user features. 
    \item PETER~\cite{li2021personalized} is a Transformer-based method to generate personalized explanations. It takes user and item IDs into consideration when generating explanations. 
\end{itemize}

Since PETER has demonstrated its better performance by human evaluation compared to other baseline models~\cite{li2021personalized}, we only perform the human evaluation on the explanations generated by PETER and our model EmoTER.

%  \vspace{-0.5cm}
\subsection{Model Training}
When training models, we split one dataset into three parts -- the training set to train models, the validation set to fine-tune hyper-parameters, and the test set to measure the performance.
The reported results are averaged on five random data splittings.
Since NETE and PETER are open-sourced, we reused the published codes and followed their default settings in training and testing.

% We train models on the training set, fine-tune the hyperparameters on the validation set, and measure the performance on the test set. The test results are averaged on the 5 data splits. Since the code of NETE and PETER is publicly available, we reuse the codes and the other default settings when training and testing.

When training the proposed EmoTER, the word embedding size $d$ was set to 512, the dimension of the feed-forward network was set to 2048, and both the number of layers and attention heads of the decoder block were set as 2. 
The hyperparameter of language modeling task $c_1$ and the hyperparameter of emotion task $c_2$ were set to 1.0 and 1.0 respectively after performing grid search.
We set the batch size to 128 and the learning rate to 1.0.
The optimizer was SGD (stochastic gradient descent), and gradient clipping was adopted  with a threshold of 1.0.

\begin{table*}[h]
\centering
 \caption{Ablation study of multitasks settings performed on the TripAdvisor dataset.$\uparrow$ and $\downarrow$ indicate performance changes compared to EmoTER. BLEU and ROUGE are percentage values.}
%\vspace{-0.3cm}
 \label{tab:multi_task}
%  \begin{tabular}{*{13}{c}}
  \begin{tabular}{c|ccc|*{9}{c}}
 \toprule
 \toprule
 & \multicolumn{3}{c|}{Explainability} & \multicolumn{9}{c}{Text quality} \\
 \midrule
 & FMR & FCR & DIV & USR & BLEU-1&BLEU-4&R1-P& R1-R&R1-F& R2-P& R2-R&R2-F \\
%  \midrule
 \midrule
 Disable $L_{emo}$&0.87 & 0.35 $\downarrow$ & 1.53$\downarrow$ & 0.30$\downarrow$ & 24.62$\downarrow$ &4.70$\downarrow$&39.41$\uparrow$& 29.62$\downarrow$&31.56$\downarrow$& 12.98$\downarrow$& 9.41$\downarrow$&9.89$\downarrow$ \\
 Disable $L_{lm}$&0.88$\uparrow$ & 0.31$\downarrow$ & 1.64$\downarrow$ & 0.20$\downarrow$ & 25.06$\uparrow$&4.76$\downarrow$&39.15$\downarrow$& 29.87$\downarrow$&31.58$\downarrow$& 12.71$\downarrow$& 9.44$\downarrow$&9.78$\downarrow$ \\
 EmoTER & 0.87&0.39 & 1.53 & 0.42 & 25.03& 4.80&39.35 & 29.95 & 31.77& 13.04& 9.56&10.03 \\
 \bottomrule
\end{tabular}
\end{table*}

\begin{table*}[h]
\setlength{\tabcolsep}{4pt}
\centering
 \caption{Ablation study of intensity performed on the TripAdvisor dataset.$\uparrow$ and $\downarrow$ indicate performance change compared to EmoTER. BLEU and ROUGE are percentage values.}
%\vspace{-0.3cm}
 \label{tab:intensity}
%  \begin{tabular}{*{13}{c}}
   \begin{tabular}{c|ccc|*{9}{c}}
 \toprule
 \toprule
 & \multicolumn{3}{c|}{Explainability} & \multicolumn{9}{c}{Text quality} \\
 \midrule
 & FMR & FCR & DIV & USR & BLEU-1&BLEU-4&R1-P& R1-R&R1-F& R2-P& R2-R&R2-F \\
 \midrule
%  \midrule
intensity=0.5 &0.87 & 0.39 & 1.57$\downarrow$ & 0.43$\uparrow$ & 25.72$\uparrow$&4.90$\uparrow$&38.56$\downarrow$& 30.32$\uparrow$&31.68$\downarrow$& 12.51$\downarrow$& 9.64$\uparrow$&9.87$\downarrow$ \\
intensity=2 &0.86$\downarrow$ & 0.37$\downarrow$ & 1.62$\downarrow$ & 0.33$\downarrow$ & 25.30$\uparrow$&4.83$\uparrow$&39.00$\downarrow$& 30.00$\uparrow$&31.65$\downarrow$& 12.69$\downarrow$& 9.52$\downarrow$&9.87$\downarrow$ \\
 EmoTER (intensity=1)& 0.87&0.39 & 1.53 & 0.42 & 25.03& 4.80&39.35 & 29.95 & 31.77& 13.04& 9.56&10.03 \\
 \bottomrule
\end{tabular}
%\vspace{-0.3cm}
\end{table*}

% \vspace{-0.3cm}
\section{Result and Analysis}
This section presents automatic and human evaluation results of EmoTER's performance.
\subsection{Automatic Evaluation on Explanation}
We compared EmoTER with existing state-of-the-art explanation generation models from the perspectives of the robustness of emotion distribution, explainability, and text quality.

\noindent\textbf{Explanations with emotion fairness and robustness} To illustrate that EmoTER is capable of reducing emotion biases in recommendation explanations, we qualitatively evaluated its debiasing performance by comparing emotion distributions in explanations generated by PETER and by the proposed EmoTER on the TripAdvisor, Amazon, and Yelp datasets.
We adopted two widely-used emotion detection models to recognize the emotion of generated explanations: (1) Text2emotion~\cite{Ni}, a lightweight Python library; and (2) GoEmotions~\cite{Demszky}.
Given an explanation, Text2emotion returns an emotion category among happy, angry, surprise, sad, fear and a strength score ranging from 0 to 1.
If the score was smaller than 0.2, we considered the explanation to be neutral.
GoEmotions is one of the state-of-the-art BERT-based model emotion detectors.

Tables~\ref{tab:Text2Emotion} and \ref{tab:Goemotion} show emotion detection results by Text2emotion and GoEmotion on the TripedAdvisor dataset, where the detected ``happy'' dominates all other emotions.
We found that the emotion distribution generated by EmoTER is more consistent with the ground-truth when compared to PETER.
Specifically, EmoTER reduces distribution biases on nine of ten detected emotions by Text2emotion and GoEmotion. 
We also observed similar debiasing performance of EmoTER on the Amazon and Yelp datasets.

% the emotion distribution in ground truth, explanations generated by PETER and emotion-aware explanations generated by our model on the three datasets shown in Table~\ref{tab:dataset}.

% As we show in Tables~\ref{tab:Text2Emotion} and \ref{tab:Goemotion}, the ground-truth recommendation explanation datasets suffer from huge emotion biases.
% For example, ``happy'' dominates the emotion distribution.
% Emotion detection results of Text2emotion and GoEmotion on the TripedAdvisor dataset are shown in Table~\ref{tab:Text2Emotion} and Table~\ref{tab:Goemotion} respectively.
% We found that the emotion distribution generated by EmoTER is much closer to the distribution of ground-truth compared to PETER+.
% % We could see that the emotion explanations generated by PETER which is the state-of-the-art model without emotion.
% Table~\ref{tab:Text2Emotion} and Table~\ref{tab:Goemotion} shows EmoTER reduces emotion biases by 21.9\%, 10.2\%, 37.8\%, 10.4\%, 30.5\%, 25.4\% and 10.0\%, 13.3\%, 6.3\%, 38.9\%, 3.3\%, 14.0\% respectively in ``happy'', ``angry'', ``surprise'', ``sad'', ``fear'' and ``neutral''.
% Thus, we found our model had a significant debiasing effect on all emotion categories. 

\noindent\textbf{Explainability} In terms of explainability, EmoTER is very competitive and close to the best performing values regarding the FMR metric. EmoTER beats the other three state-of-the-art models by a large margin on FCR metric which considers whether explanations contain all the features in the ground-truth.
EmoTER achieves much better performance on explainability even compared to pre-trained models (e.g., ACMLM) that learn large-scale external knowledge. 
Lower DIV means higher diversity. It is not surprising that ACMLM has the highest diversity. However, EmoTER also achieves a competitive score.

% \noindent\textbf{Explainability} In terms of explainability, EmoTER beats the other three state-of-the-art models by a large margin on FCR metric which considers whether explanations contain all the features in the ground-truth. In particular, ACMLM is a pre-trained model which learns large-scale external knowledge. EmoTER achieves much better performance on explainability even compared with pre-trained models. Lower DIV means higher diversity. It is not surprising that ACMLM has the highest diversity. However, EmoTER also achieves a competitive score.
% Regarding the FMR metric, EmoTER is also very competitive and close to the best score. 

\noindent\textbf{Text quality} We also compared the text quality of generated text by EmoTER with three state-of-the-art baselines on three datasets (see the text quality columns in Table~\ref{tab:quantitative} for details).
We used unique sentence ratio (USR), an important metric to evaluate diversity on the sentence level, to measure text diversity.
EmoTER performs much better than PETER on USR and slightly worse than NETE.
It is reasonable that NETE is more likely to generates diverse explanations as it is a template-like approach.
ACMLM achieves the highest USR because it is a well-tuned model trained on a large-scale dataset.
We also use BLUE-1, BLUE-4, ROUGE-1 (R1 column in Table~\ref{tab:quantitative}), and ROUGE (R2 column in Table~\ref{tab:quantitative}) to evaluate text quality. 
EmoTER consistently outperforms baselines by a large margin on BLEU and ROUGE metrics, verifying our model's ability to generate high-quality explanations.

%\vspace{-0.3cm}
\noindent\textbf{Ablation Study} As an additional checkpoint and evaluation, we conducted an ablation study on the TripAdvisor dataset to show the effectiveness of our multi-task design shown in Table~\ref{tab:multi_task} and hyper-parameter of intensity shown in Table~\ref{tab:intensity}.
First, we launched an experiment without emotion classification loss by setting $c_1$ in Equation~\ref{equ:loss_total} to be 0.
We found that disabling the emotion classification task causes USR to drop by nearly 25\%, and performances of almost all text quality metrics decrease sharply.
Similarly, we disabled language modeling loss by setting $c_2$ to be 0.
It not only degraded the explainability performance but also caused USR to drop nearly 50\%.
These two experiments confirm EmoTER's effectiveness in generating emotional and high-quality explanations. 
In addition, we changed the intensity strength to generate emotional explanations with different emotion intensities shown in Table~\ref{tab:intensity}.
We found that the explanations maintain high performance of explainability and text quality when intensity was small. If emotion intensity becomes too strong, it may cause performance of USR and DIV to drop dramatically.

\begin{table}[t]
    \setlength{\tabcolsep}{4pt}
    \centering 
    \caption{Qualitative examples from three datasets}
    \label{tab:emotion_case}
    %\vspace{-0.3cm}
    \begin{tabular}{p{0.14\linewidth}p{0.3\linewidth}p{0.2\linewidth}p{0.3\linewidth}}
    \toprule
    \toprule
    &  TripAdvisor  & Yelp & Amazon\\
    \midrule
    Ground Truth & Beautiful lobby and nice bar. & The wait is long & Don't waste your time on this loser.\\
    PETER & The lobby was very nice and the rooms were very comfortable & Wait for food. & If you are a fan of the original, you'll love this film.  \\
    EmoTER & The lobby is impressive and the rooms are spacious&Don't worth the wait. & If you are a fan of the films, you will be disappointed. \\
    \bottomrule
    \end{tabular}
    %\vspace{-0.3cm}
\end{table}

\begin{figure}[t]
  \centering
  
  \begin{minipage}[b]{0.5\textwidth}
    % \vspace{-1.5cm}
    \centering
    \includegraphics[width=0.6\linewidth]{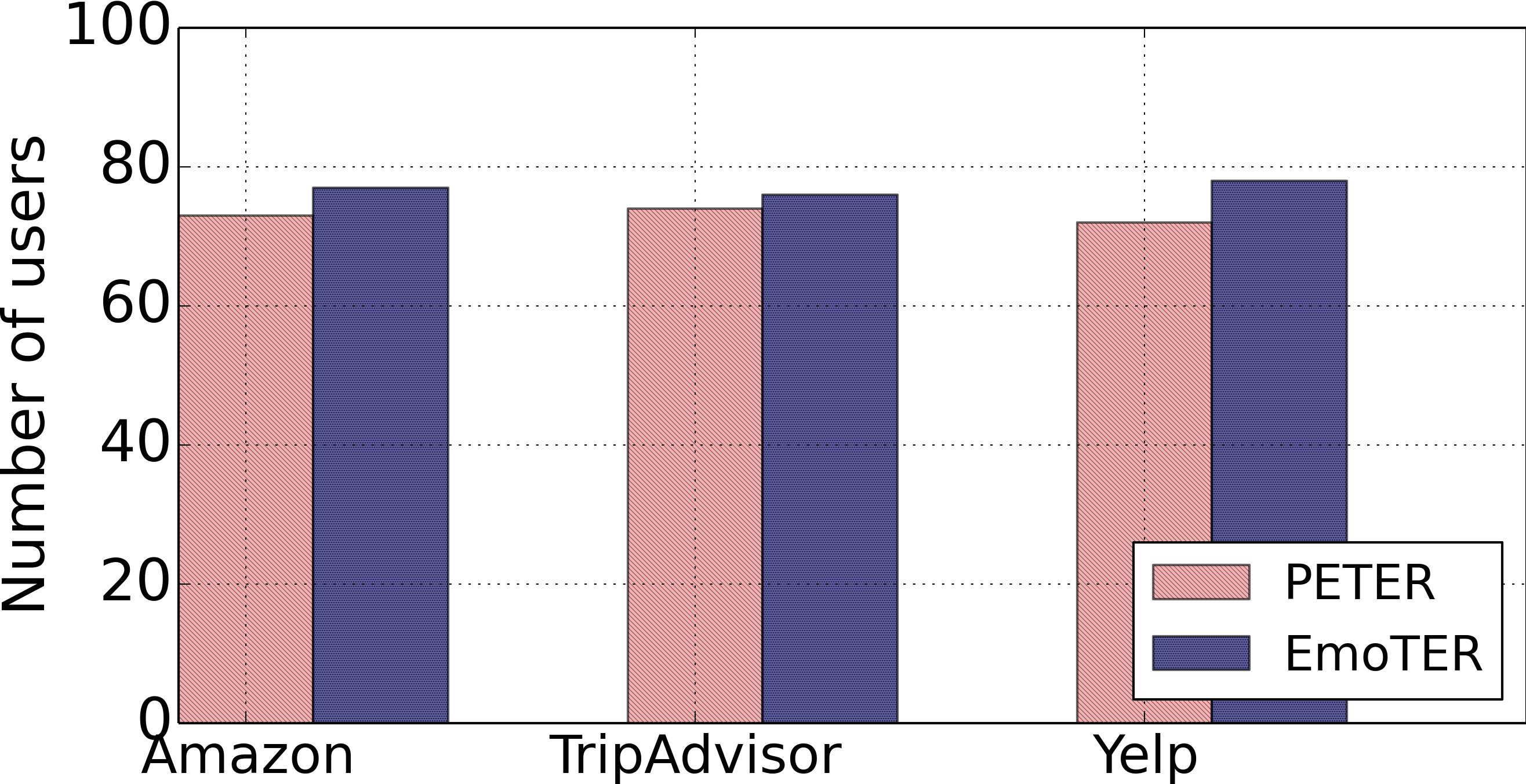}
    \caption{Human evaluation result about emotion robustness on generated explanations on three benchmark datasets. The t-test result demonstrated that EmoTER outperformed PETER significantly in generating robust explanations ($p<0.05$).
    % We also perform student t-test. The result is statistically significant which shows EmoTER generates more robust explanations
    }
  \label{fig:robust_rating}
  \end{minipage}
  \end{figure}
  
\subsection{Human Evaluation on Explanations}

In addition to automatic metrics, we also evaluate human perceptions of emotion-aware explanations. We conducted one Amazon MTurk user study to validate that our model EmoTER increases the robustness and fairness of emotion-aware explanation.

% In addition to automatic metrics, we also evaluate human perceptions of emotion-aware explanations. We conducted two different user studies to validate that our model EmoTER increases the robustness and fairness of emotion-aware explanation without losing much of the informativeness, friendliness, credibility, quality, human-likeness, and even behaves better in certain dimensions.

% So far, we have shown the technical effectiveness of EmoTER based on automatic evaluation metrics, but this presents only one perspective of this approach. The other perspective is about the end-users' perceptions of the emotion-aware explanations. In this part, we show two different user study results to validate that our model EmoTER increases the robustness and fairness of emotion-aware explanation without losing much of the informativeness, friendliness, credibility, quality, human-likeness, and even behaves better in certain dimensions.

\noindent\textbf{Explanations with emotion fairness and robustness} We show qualitative examples from three benchmark datasets in Table~\ref{tab:emotion_case}.
An example of TripAdvisor dataset shows that although these three explanations represent the same happy emotion, our model EmoTER generates more diverse and unique vocabularies.
An example of Yelp dataset shows that EmoTER generates explanations with the same negative emotion as ground-truth while PETER generates positive emotion.
An example of Amazon dataset also shows that our model EmoTER generates a more robust explanation with consistent emotion with input. This indicates a higher level of {\em fairness} as it relates to being representative of the underlying distribution of ground truth emotions (disparate impact notion of fairness).

Figure~\ref{fig:robust_rating} shows the human evaluation results of explanations with emotion fairness and robustness. More users think explanations generated by our model EmoTER are emotionally closer to the ground-truth than explanations generated by the current state-of-the-art model PETER. We can see that our model EmoTER consistently outperforms PETER in three benchmark datasets in terms of robustness and fairness. We perform student t-test among the human evaluation result and obtain significant result ($t=4.90, p=0.008$).

\section{Conclusion}

Providing explanations that are appropriately emotion-induced could help users of recommender systems understand and trust the recommendations better, but existing approaches have failed to incorporate a wide range of emotions in unbiased and naturalistic manner. In this paper, we 
%first address the emotion bias issues on three benchmark datasets and current state-of-the-art models for recommendation explanation and 
proposed an emotion-aware generation framework to generate robust and fair explanations. To the best of our knowledge, this is the first work to explore robust, fair, emotion-aware explanation generation in explainable recommendations. We presented EmoTER (Emotion-aware Transformer for Explainable Recommendation), a novel multi-head Transformer architecture, which enhances the emotion of the explanations with a multi-task learning approach for personalized explanation generation. Experiments on three widely-used benchmark datasets with multiple evaluation metrics demonstrated that EmoTER consistently outperforms existing state-of-the-art explanation generation models in terms of robustness of emotion distribution, text quality, and explainability. 

% Crowd-sourced experiments showed that EmoTER increased the robustness of emotion-aware explanation without losing much of the informativeness, friendliness, credibility, quality, human-likeness, and even behaves better in certain dimensions.

The next steps in this research will involve paying attention to the possible dataset biases and further exploring fairer methods to generate explanations for recommendation systems. Moreover, we will extend our framework to emotion-aware multi-modal explanations and develop our solution for explainable conversational recommendations, which significantly needs the emotional elements to enhance the user engagement with various informational systems, including conversational agents. 

% \newpage

\bibliographystyle{ACM-Reference-Format}
\bibliography{main}


\begin{thebibliography}{46}

%%% the Web address.

\ifx \showCODEN    \undefined \def \showCODEN     #1{\unskip}     \fi
\ifx \showDOI      \undefined \def \showDOI       #1{#1}\fi
\ifx \showISBNx    \undefined \def \showISBNx     #1{\unskip}     \fi
\ifx \showISBNxiii \undefined \def \showISBNxiii  #1{\unskip}     \fi
\ifx \showISSN     \undefined \def \showISSN      #1{\unskip}     \fi
\ifx \showLCCN     \undefined \def \showLCCN      #1{\unskip}     \fi
\ifx \shownote     \undefined \def \shownote      #1{#1}          \fi
\ifx \showarticletitle \undefined \def \showarticletitle #1{#1}   \fi
\ifx \showURL      \undefined \def \showURL       {\relax}        \fi
% invisible to TeX
\providecommand\bibfield[2]{#2}
\providecommand\bibinfo[2]{#2}
\providecommand\natexlab[1]{#1}
\providecommand\showeprint[2][]{arXiv:#2}

\bibitem[\protect\citeauthoryear{Asghar, Poupart, Hoey, Jiang, and Mou}{Asghar
  et~al\mbox{.}}{2017}]%
        {asghar2017affective}
\bibfield{author}{\bibinfo{person}{Nabiha Asghar}, \bibinfo{person}{Pascal
  Poupart}, \bibinfo{person}{Jesse Hoey}, \bibinfo{person}{Xin Jiang}, {and}
  \bibinfo{person}{Lili Mou}.} \bibinfo{year}{2017}\natexlab{}.
\newblock \bibinfo{title}{Affective Neural Response Generation}.
\newblock
\newblock
\showeprint[arxiv]{1709.03968}~[cs.CL]


\bibitem[\protect\citeauthoryear{Balog and Radlinski}{Balog and
  Radlinski}{2020}]%
        {Google}
\bibfield{author}{\bibinfo{person}{Krisztian Balog} {and}
  \bibinfo{person}{Filip Radlinski}.} \bibinfo{year}{2020}\natexlab{}.
\newblock \showarticletitle{{Measuring Recommendation Explanation Quality: The
  Conflicting Goals of Explanations}}.
\newblock  (\bibinfo{year}{2020}), \bibinfo{pages}{10}.
\newblock
\showISBNx{9781450380164}
\urldef\tempurl%
\url{https://doi.org/10.1145/3397271.3401032}
\showDOI{\tempurl}


\bibitem[\protect\citeauthoryear{Banerjee and Lavie}{Banerjee and
  Lavie}{2005}]%
        {banerjee2005meteor}
\bibfield{author}{\bibinfo{person}{Satanjeev Banerjee} {and}
  \bibinfo{person}{Alon Lavie}.} \bibinfo{year}{2005}\natexlab{}.
\newblock \showarticletitle{METEOR: An automatic metric for MT evaluation with
  improved correlation with human judgments}. In
  \bibinfo{booktitle}{\emph{Proceedings of the acl workshop on intrinsic and
  extrinsic evaluation measures for machine translation and/or summarization}}.
  \bibinfo{pages}{65--72}.
\newblock


\bibitem[\protect\citeauthoryear{Bras, Swayamdipta, Bhagavatula, Zellers,
  Peters, Sabharwal, and Choi}{Bras et~al\mbox{.}}{2020}]%
        {pmlr-v119-bras20a}
\bibfield{author}{\bibinfo{person}{Ronan~Le Bras}, \bibinfo{person}{Swabha
  Swayamdipta}, \bibinfo{person}{Chandra Bhagavatula}, \bibinfo{person}{Rowan
  Zellers}, \bibinfo{person}{Matthew Peters}, \bibinfo{person}{Ashish
  Sabharwal}, {and} \bibinfo{person}{Yejin Choi}.}
  \bibinfo{year}{2020}\natexlab{}.
\newblock \showarticletitle{Adversarial Filters of Dataset Biases}. In
  \bibinfo{booktitle}{\emph{Proceedings of the 37th International Conference on
  Machine Learning}} \emph{(\bibinfo{series}{Proceedings of Machine Learning
  Research}, Vol.~\bibinfo{volume}{119})},
  \bibfield{editor}{\bibinfo{person}{Hal~Daumé III} {and}
  \bibinfo{person}{Aarti Singh}} (Eds.). \bibinfo{publisher}{PMLR},
  \bibinfo{pages}{1078--1088}.
\newblock
\urldef\tempurl%
\url{https://proceedings.mlr.press/v119/bras20a.html}
\showURL{%
\tempurl}


\bibitem[\protect\citeauthoryear{Chen, Chen, Shi, and Zhang}{Chen
  et~al\mbox{.}}{2021a}]%
        {chen2021generate}
\bibfield{author}{\bibinfo{person}{Hanxiong Chen}, \bibinfo{person}{Xu Chen},
  \bibinfo{person}{Shaoyun Shi}, {and} \bibinfo{person}{Yongfeng Zhang}.}
  \bibinfo{year}{2021}\natexlab{a}.
\newblock \showarticletitle{Generate natural language explanations for
  recommendation}.
\newblock \bibinfo{journal}{\emph{arXiv preprint arXiv:2101.03392}}
  (\bibinfo{year}{2021}).
\newblock


\bibitem[\protect\citeauthoryear{Chen, Chen, Shi, and Zhang}{Chen
  et~al\mbox{.}}{2021b}]%
        {DBLP:journals/corr/abs-2101-03392}
\bibfield{author}{\bibinfo{person}{Hanxiong Chen}, \bibinfo{person}{Xu Chen},
  \bibinfo{person}{Shaoyun Shi}, {and} \bibinfo{person}{Yongfeng Zhang}.}
  \bibinfo{year}{2021}\natexlab{b}.
\newblock \showarticletitle{Generate Natural Language Explanations for
  Recommendation}.
\newblock \bibinfo{journal}{\emph{CoRR}}  \bibinfo{volume}{abs/2101.03392}
  (\bibinfo{year}{2021}).
\newblock
\showeprint[arxiv]{2101.03392}
\urldef\tempurl%
\url{https://arxiv.org/abs/2101.03392}
\showURL{%
\tempurl}


\bibitem[\protect\citeauthoryear{Chen, Chen, Xu, Zhang, Cao, Qin, and Zha}{Chen
  et~al\mbox{.}}{2019}]%
        {Chen_Zhang_Qin_2019}
\bibfield{author}{\bibinfo{person}{Xu Chen}, \bibinfo{person}{Hanxiong Chen},
  \bibinfo{person}{Hongteng Xu}, \bibinfo{person}{Yongfeng Zhang},
  \bibinfo{person}{Yixin Cao}, \bibinfo{person}{Zheng Qin}, {and}
  \bibinfo{person}{Hongyuan Zha}.} \bibinfo{year}{2019}\natexlab{}.
\newblock \showarticletitle{Personalized fashion recommendation with visual
  explanations based on multimodal attention network: Towards visually
  explainable recommendation}. In \bibinfo{booktitle}{\emph{Proceedings of the
  42nd International ACM SIGIR Conference on Research and Development in
  Information Retrieval}}. \bibinfo{pages}{765--774}.
\newblock


\bibitem[\protect\citeauthoryear{Chen, Qin, Zhang, and Xu}{Chen
  et~al\mbox{.}}{2016}]%
        {chen2016learning}
\bibfield{author}{\bibinfo{person}{Xu Chen}, \bibinfo{person}{Zheng Qin},
  \bibinfo{person}{Yongfeng Zhang}, {and} \bibinfo{person}{Tao Xu}.}
  \bibinfo{year}{2016}\natexlab{}.
\newblock \showarticletitle{Learning to rank features for recommendation over
  multiple categories}. In \bibinfo{booktitle}{\emph{Proceedings of the 39th
  International ACM SIGIR conference on Research and Development in Information
  Retrieval}}. ACM, \bibinfo{pages}{305--314}.
\newblock


\bibitem[\protect\citeauthoryear{Colombo, Witon, Modi, Kennedy, and
  Kapadia}{Colombo et~al\mbox{.}}{2019}]%
        {Colombo2019AffectDrivenDG}
\bibfield{author}{\bibinfo{person}{Pierre Colombo}, \bibinfo{person}{Wojciech
  Witon}, \bibinfo{person}{Ashutosh Modi}, \bibinfo{person}{J. Kennedy}, {and}
  \bibinfo{person}{Mubbasir Kapadia}.} \bibinfo{year}{2019}\natexlab{}.
\newblock \showarticletitle{Affect-Driven Dialog Generation}. In
  \bibinfo{booktitle}{\emph{NAACL}}.
\newblock


\bibitem[\protect\citeauthoryear{Demszky, Movshovitz-Attias, Ko, Cowen, Nemade,
  and Ravi}{Demszky et~al\mbox{.}}{[n.\,d.]}]%
        {Demszky}
\bibfield{author}{\bibinfo{person}{Dorottya Demszky}, \bibinfo{person}{Dana
  Movshovitz-Attias}, \bibinfo{person}{Jeongwoo Ko}, \bibinfo{person}{Alan
  Cowen}, \bibinfo{person}{Gaurav Nemade}, {and} \bibinfo{person}{Sujith
  Ravi}.} \bibinfo{year}{[n.\,d.]}\natexlab{}.
\newblock \showarticletitle{{GoEmotions: A Dataset of Fine-Grained Emotions}}.
\newblock  (\bibinfo{year}{[n.\,d.]}).
\newblock
\showeprint[arxiv]{2005.00547v2}
\urldef\tempurl%
\url{https://github.com/dewarim/}
\showURL{%
\tempurl}


\bibitem[\protect\citeauthoryear{Devlin, Chang, Lee, and Toutanova}{Devlin
  et~al\mbox{.}}{2019}]%
        {devlin2019bert}
\bibfield{author}{\bibinfo{person}{Jacob Devlin}, \bibinfo{person}{Ming-Wei
  Chang}, \bibinfo{person}{Kenton Lee}, {and} \bibinfo{person}{Kristina
  Toutanova}.} \bibinfo{year}{2019}\natexlab{}.
\newblock \bibinfo{title}{BERT: Pre-training of Deep Bidirectional Transformers
  for Language Understanding}.
\newblock
\newblock
\showeprint[arxiv]{1810.04805}~[cs.CL]


\bibitem[\protect\citeauthoryear{Fiehler}{Fiehler}{2002}]%
        {fiehler2002emotions}
\bibfield{author}{\bibinfo{person}{Reinhard Fiehler}.}
  \bibinfo{year}{2002}\natexlab{}.
\newblock \showarticletitle{How to do emotions with words: Emotionality in
  conversations}.
\newblock In \bibinfo{booktitle}{\emph{The verbal communication of emotions}}.
  \bibinfo{publisher}{Psychology Press}, \bibinfo{pages}{87--114}.
\newblock


\bibitem[\protect\citeauthoryear{Ford}{Ford}{1992}]%
        {ford1992motivating}
\bibfield{author}{\bibinfo{person}{Martin~E Ford}.}
  \bibinfo{year}{1992}\natexlab{}.
\newblock \bibinfo{booktitle}{\emph{Motivating humans: Goals, emotions, and
  personal agency beliefs}}.
\newblock \bibinfo{publisher}{Sage}.
\newblock


\bibitem[\protect\citeauthoryear{Ghosh, Chollet, Laksana, Morency, and
  Scherer}{Ghosh et~al\mbox{.}}{2017}]%
        {Ghosh}
\bibfield{author}{\bibinfo{person}{Sayan Ghosh}, \bibinfo{person}{Mathieu
  Chollet}, \bibinfo{person}{Eugene Laksana}, \bibinfo{person}{Louis-Philippe
  Morency}, {and} \bibinfo{person}{Stefan Scherer}.}
  \bibinfo{year}{2017}\natexlab{}.
\newblock \showarticletitle{Affect-LM: A Neural Language Model for Customizable
  Affective Text Generation}. In \bibinfo{booktitle}{\emph{Proceedings of the
  55th Annual Meeting of the Association for Computational Linguistics (Volume
  1: Long Papers)}}. \bibinfo{pages}{634--642}.
\newblock


\bibitem[\protect\citeauthoryear{Goswamy, Singh, Barkati, and Modi}{Goswamy
  et~al\mbox{.}}{2020}]%
        {Singh}
\bibfield{author}{\bibinfo{person}{Tushar Goswamy}, \bibinfo{person}{Ishika
  Singh}, \bibinfo{person}{Ahsan Barkati}, {and} \bibinfo{person}{Ashutosh
  Modi}.} \bibinfo{year}{2020}\natexlab{}.
\newblock \showarticletitle{Adapting a language model for controlled affective
  text generation}. In \bibinfo{booktitle}{\emph{Proceedings of the 28th
  International Conference on Computational Linguistics}}.
  \bibinfo{pages}{2787--2801}.
\newblock


\bibitem[\protect\citeauthoryear{Herlocker, Konstan, and Riedl}{Herlocker
  et~al\mbox{.}}{2000}]%
        {herlocker2000explaining}
\bibfield{author}{\bibinfo{person}{Jonathan~L Herlocker},
  \bibinfo{person}{Joseph~A Konstan}, {and} \bibinfo{person}{John Riedl}.}
  \bibinfo{year}{2000}\natexlab{}.
\newblock \showarticletitle{Explaining collaborative filtering
  recommendations}. In \bibinfo{booktitle}{\emph{Proceedings of the 2000 ACM
  conference on Computer supported cooperative work}}. ACM,
  \bibinfo{pages}{241--250}.
\newblock


\bibitem[\protect\citeauthoryear{Hochreiter and Schmidhuber}{Hochreiter and
  Schmidhuber}{1997}]%
        {6795963}
\bibfield{author}{\bibinfo{person}{Sepp Hochreiter} {and}
  \bibinfo{person}{Jürgen Schmidhuber}.} \bibinfo{year}{1997}\natexlab{}.
\newblock \showarticletitle{Long Short-Term Memory}.
\newblock \bibinfo{journal}{\emph{Neural Computation}} \bibinfo{volume}{9},
  \bibinfo{number}{8} (\bibinfo{year}{1997}), \bibinfo{pages}{1735--1780}.
\newblock
\urldef\tempurl%
\url{https://doi.org/10.1162/neco.1997.9.8.1735}
\showDOI{\tempurl}


\bibitem[\protect\citeauthoryear{Huang, Lee, Ma, Chen, Yu, and Chen}{Huang
  et~al\mbox{.}}{[n.\,d.]}]%
        {Huang}
\bibfield{author}{\bibinfo{person}{Yen-Hao Huang}, \bibinfo{person}{Ssu-Rui
  Lee}, \bibinfo{person}{Mau-Yun Ma}, \bibinfo{person}{Yi-Hsin Chen},
  \bibinfo{person}{Ya-Wen Yu}, {and} \bibinfo{person}{Yi-Shin Chen}.}
  \bibinfo{year}{[n.\,d.]}\natexlab{}.
\newblock \showarticletitle{{EmotionX-IDEA: Emotion BERT-an Affectional Model
  for Conversation}}.
\newblock  (\bibinfo{year}{[n.\,d.]}).
\newblock
\showeprint[arxiv]{1908.06264v1}
\urldef\tempurl%
\url{http://nlp.mathcs.emory.edu}
\showURL{%
\tempurl}


\bibitem[\protect\citeauthoryear{Keshtkar and Inkpen}{Keshtkar and
  Inkpen}{2011}]%
        {10.1007/978-3-642-24571-8_2}
\bibfield{author}{\bibinfo{person}{Fazel Keshtkar} {and} \bibinfo{person}{Diana
  Inkpen}.} \bibinfo{year}{2011}\natexlab{}.
\newblock \showarticletitle{A Pattern-Based Model for Generating Text to
  Express Emotion}. In \bibinfo{booktitle}{\emph{Affective Computing and
  Intelligent Interaction}}, \bibfield{editor}{\bibinfo{person}{Sidney
  D'Mello}, \bibinfo{person}{Arthur Graesser}, \bibinfo{person}{Bj{\"o}rn
  Schuller}, {and} \bibinfo{person}{Jean-Claude Martin}} (Eds.).
  \bibinfo{publisher}{Springer Berlin Heidelberg}, \bibinfo{address}{Berlin,
  Heidelberg}, \bibinfo{pages}{11--21}.
\newblock
\showISBNx{978-3-642-24571-8}


\bibitem[\protect\citeauthoryear{Kim, Bak, and Oh}{Kim et~al\mbox{.}}{2012}]%
        {kim2012you}
\bibfield{author}{\bibinfo{person}{Suin Kim}, \bibinfo{person}{JinYeong Bak},
  {and} \bibinfo{person}{Alice~Haeyun Oh}.} \bibinfo{year}{2012}\natexlab{}.
\newblock \showarticletitle{Do you feel what i feel? social aspects of emotions
  in twitter conversations}. In \bibinfo{booktitle}{\emph{Sixth International
  AAAI Conference on Weblogs and Social Media}}.
\newblock


\bibitem[\protect\citeauthoryear{K{\"o}per, Kim, and Klinger}{K{\"o}per
  et~al\mbox{.}}{2017}]%
        {koper-etal-2017-ims}
\bibfield{author}{\bibinfo{person}{Maximilian K{\"o}per},
  \bibinfo{person}{Evgeny Kim}, {and} \bibinfo{person}{Roman Klinger}.}
  \bibinfo{year}{2017}\natexlab{}.
\newblock \showarticletitle{{IMS} at {E}mo{I}nt-2017: Emotion Intensity
  Prediction with Affective Norms, Automatically Extended Resources and Deep
  Learning}. In \bibinfo{booktitle}{\emph{Proceedings of the 8th Workshop on
  Computational Approaches to Subjectivity, Sentiment and Social Media
  Analysis}}. \bibinfo{publisher}{Association for Computational Linguistics},
  \bibinfo{address}{Copenhagen, Denmark}, \bibinfo{pages}{50--57}.
\newblock
\urldef\tempurl%
\url{https://doi.org/10.18653/v1/W17-5206}
\showDOI{\tempurl}


\bibitem[\protect\citeauthoryear{Lahoti, Beutel, Chen, Lee, Prost, Thain, Wang,
  and Chi}{Lahoti et~al\mbox{.}}{2020}]%
        {lahoti2020fairness}
\bibfield{author}{\bibinfo{person}{Preethi Lahoti}, \bibinfo{person}{Alex
  Beutel}, \bibinfo{person}{Jilin Chen}, \bibinfo{person}{Kang Lee},
  \bibinfo{person}{Flavien Prost}, \bibinfo{person}{Nithum Thain},
  \bibinfo{person}{Xuezhi Wang}, {and} \bibinfo{person}{Ed~H Chi}.}
  \bibinfo{year}{2020}\natexlab{}.
\newblock \showarticletitle{Fairness without demographics through adversarially
  reweighted learning}.
\newblock \bibinfo{journal}{\emph{arXiv preprint arXiv:2006.13114}}
  (\bibinfo{year}{2020}).
\newblock


\bibitem[\protect\citeauthoryear{Lewis, Stenetorp, and Riedel}{Lewis
  et~al\mbox{.}}{2021}]%
        {lewis-etal-2021-question}
\bibfield{author}{\bibinfo{person}{Patrick Lewis}, \bibinfo{person}{Pontus
  Stenetorp}, {and} \bibinfo{person}{Sebastian Riedel}.}
  \bibinfo{year}{2021}\natexlab{}.
\newblock \showarticletitle{Question and Answer Test-Train Overlap in
  Open-Domain Question Answering Datasets}. In
  \bibinfo{booktitle}{\emph{Proceedings of the 16th Conference of the European
  Chapter of the Association for Computational Linguistics: Main Volume}}.
  \bibinfo{publisher}{Association for Computational Linguistics},
  \bibinfo{address}{Online}, \bibinfo{pages}{1000--1008}.
\newblock
\urldef\tempurl%
\url{https://doi.org/10.18653/v1/2021.eacl-main.86}
\showDOI{\tempurl}


\bibitem[\protect\citeauthoryear{Li, Zhang, and Chen}{Li et~al\mbox{.}}{2020}]%
        {Li2020}
\bibfield{author}{\bibinfo{person}{Lei Li}, \bibinfo{person}{Yongfeng Zhang},
  {and} \bibinfo{person}{Li Chen}.} \bibinfo{year}{2020}\natexlab{}.
\newblock \showarticletitle{{Generate Neural Template Expla-nations for
  Recommendation}}.
\newblock  (\bibinfo{year}{2020}).
\newblock
\showISBNx{9781450368599}
\urldef\tempurl%
\url{https://doi.org/10.1145/3340531.3411992}
\showDOI{\tempurl}


\bibitem[\protect\citeauthoryear{Li, Zhang, and Chen}{Li
  et~al\mbox{.}}{2021a}]%
        {li2021extra}
\bibfield{author}{\bibinfo{person}{Lei Li}, \bibinfo{person}{Yongfeng Zhang},
  {and} \bibinfo{person}{Li Chen}.} \bibinfo{year}{2021}\natexlab{a}.
\newblock \bibinfo{title}{EXTRA: Explanation Ranking Datasets for Explainable
  Recommendation}.
\newblock
\newblock
\showeprint[arxiv]{2102.10315}~[cs.IR]


\bibitem[\protect\citeauthoryear{Li, Zhang, and Chen}{Li
  et~al\mbox{.}}{2021b}]%
        {li2021personalized}
\bibfield{author}{\bibinfo{person}{Lei Li}, \bibinfo{person}{Yongfeng Zhang},
  {and} \bibinfo{person}{Li Chen}.} \bibinfo{year}{2021}\natexlab{b}.
\newblock \showarticletitle{Personalized Transformer for Explainable
  Recommendation}.
\newblock \bibinfo{journal}{\emph{ACL}} (\bibinfo{year}{2021}).
\newblock


\bibitem[\protect\citeauthoryear{Lin}{Lin}{2004}]%
        {lin-2004-rouge}
\bibfield{author}{\bibinfo{person}{Chin-Yew Lin}.}
  \bibinfo{year}{2004}\natexlab{}.
\newblock \showarticletitle{{ROUGE}: A Package for Automatic Evaluation of
  Summaries}. In \bibinfo{booktitle}{\emph{Text Summarization Branches Out}}.
  \bibinfo{publisher}{Association for Computational Linguistics},
  \bibinfo{address}{Barcelona, Spain}, \bibinfo{pages}{74--81}.
\newblock
\urldef\tempurl%
\url{https://www.aclweb.org/anthology/W04-1013}
\showURL{%
\tempurl}


\bibitem[\protect\citeauthoryear{Liu, Haghgoo, Chen, Raghunathan, Koh, Sagawa,
  Liang, and Finn}{Liu et~al\mbox{.}}{2021}]%
        {pmlr-v139-liu21f}
\bibfield{author}{\bibinfo{person}{Evan~Z Liu}, \bibinfo{person}{Behzad
  Haghgoo}, \bibinfo{person}{Annie~S Chen}, \bibinfo{person}{Aditi
  Raghunathan}, \bibinfo{person}{Pang~Wei Koh}, \bibinfo{person}{Shiori
  Sagawa}, \bibinfo{person}{Percy Liang}, {and} \bibinfo{person}{Chelsea
  Finn}.} \bibinfo{year}{2021}\natexlab{}.
\newblock \showarticletitle{Just Train Twice: Improving Group Robustness
  without Training Group Information}. In \bibinfo{booktitle}{\emph{Proceedings
  of the 38th International Conference on Machine Learning}}
  \emph{(\bibinfo{series}{Proceedings of Machine Learning Research},
  Vol.~\bibinfo{volume}{139})}, \bibfield{editor}{\bibinfo{person}{Marina
  Meila} {and} \bibinfo{person}{Tong Zhang}} (Eds.). \bibinfo{publisher}{PMLR},
  \bibinfo{pages}{6781--6792}.
\newblock
\urldef\tempurl%
\url{https://proceedings.mlr.press/v139/liu21f.html}
\showURL{%
\tempurl}


\bibitem[\protect\citeauthoryear{McCoy, Pavlick, and Linzen}{McCoy
  et~al\mbox{.}}{2019}]%
        {mccoy-etal-2019-right}
\bibfield{author}{\bibinfo{person}{Tom McCoy}, \bibinfo{person}{Ellie Pavlick},
  {and} \bibinfo{person}{Tal Linzen}.} \bibinfo{year}{2019}\natexlab{}.
\newblock \showarticletitle{Right for the Wrong Reasons: Diagnosing Syntactic
  Heuristics in Natural Language Inference}. In
  \bibinfo{booktitle}{\emph{Proceedings of the 57th Annual Meeting of the
  Association for Computational Linguistics}}. \bibinfo{publisher}{Association
  for Computational Linguistics}, \bibinfo{address}{Florence, Italy},
  \bibinfo{pages}{3428--3448}.
\newblock
\urldef\tempurl%
\url{https://doi.org/10.18653/v1/P19-1334}
\showDOI{\tempurl}


\bibitem[\protect\citeauthoryear{Mohammad}{Mohammad}{2018}]%
        {LREC18-AIL}
\bibfield{author}{\bibinfo{person}{Saif~M. Mohammad}.}
  \bibinfo{year}{2018}\natexlab{}.
\newblock \showarticletitle{Word Affect Intensities}. In
  \bibinfo{booktitle}{\emph{Proceedings of the 11th Edition of the Language
  Resources and Evaluation Conference (LREC-2018)}}.
  \bibinfo{address}{Miyazaki, Japan}.
\newblock


\bibitem[\protect\citeauthoryear{Nam, Cha, Ahn, Lee, and Shin}{Nam
  et~al\mbox{.}}{2020}]%
        {nam2020learning}
\bibfield{author}{\bibinfo{person}{Junhyun Nam}, \bibinfo{person}{Hyuntak Cha},
  \bibinfo{person}{Sungsoo Ahn}, \bibinfo{person}{Jaeho Lee}, {and}
  \bibinfo{person}{Jinwoo Shin}.} \bibinfo{year}{2020}\natexlab{}.
\newblock \showarticletitle{Learning from failure: Training debiased classifier
  from biased classifier}.
\newblock \bibinfo{journal}{\emph{arXiv preprint arXiv:2007.02561}}
  (\bibinfo{year}{2020}).
\newblock


\bibitem[\protect\citeauthoryear{Ni, Li, and McAuley}{Ni et~al\mbox{.}}{2019}]%
        {Ni}
\bibfield{author}{\bibinfo{person}{Jianmo Ni}, \bibinfo{person}{Jiacheng Li},
  {and} \bibinfo{person}{Julian McAuley}.} \bibinfo{year}{2019}\natexlab{}.
\newblock \showarticletitle{Justifying recommendations using distantly-labeled
  reviews and fine-grained aspects}. In \bibinfo{booktitle}{\emph{Proceedings
  of the 2019 Conference on Empirical Methods in Natural Language Processing
  and the 9th International Joint Conference on Natural Language Processing
  (EMNLP-IJCNLP)}}. \bibinfo{pages}{188--197}.
\newblock


\bibitem[\protect\citeauthoryear{Papineni, Roukos, Ward, and Zhu}{Papineni
  et~al\mbox{.}}{2002}]%
        {papineni-etal-2002-bleu}
\bibfield{author}{\bibinfo{person}{Kishore Papineni}, \bibinfo{person}{Salim
  Roukos}, \bibinfo{person}{Todd Ward}, {and} \bibinfo{person}{Wei-Jing Zhu}.}
  \bibinfo{year}{2002}\natexlab{}.
\newblock \showarticletitle{{B}leu: a Method for Automatic Evaluation of
  Machine Translation}. In \bibinfo{booktitle}{\emph{Proceedings of the 40th
  Annual Meeting of the Association for Computational Linguistics}}.
  \bibinfo{publisher}{Association for Computational Linguistics},
  \bibinfo{address}{Philadelphia, Pennsylvania, USA},
  \bibinfo{pages}{311--318}.
\newblock
\urldef\tempurl%
\url{https://doi.org/10.3115/1073083.1073135}
\showDOI{\tempurl}


\bibitem[\protect\citeauthoryear{Parkinson}{Parkinson}{1996}]%
        {parkinson1996emotions}
\bibfield{author}{\bibinfo{person}{Brian Parkinson}.}
  \bibinfo{year}{1996}\natexlab{}.
\newblock \showarticletitle{Emotions are social}.
\newblock \bibinfo{journal}{\emph{British journal of psychology}}
  \bibinfo{volume}{87}, \bibinfo{number}{4} (\bibinfo{year}{1996}),
  \bibinfo{pages}{663--683}.
\newblock


\bibitem[\protect\citeauthoryear{Plutchik}{Plutchik}{1962}]%
        {plutchik1962emotions}
\bibfield{author}{\bibinfo{person}{Robert Plutchik}.}
  \bibinfo{year}{1962}\natexlab{}.
\newblock \showarticletitle{The Emotions: Facts}.
\newblock \bibinfo{journal}{\emph{Theories and a New Model, New York}}
  (\bibinfo{year}{1962}).
\newblock


\bibitem[\protect\citeauthoryear{Plutchik}{Plutchik}{1994}]%
        {Plutchik1994ThePA}
\bibfield{author}{\bibinfo{person}{Robert Plutchik}.}
  \bibinfo{year}{1994}\natexlab{}.
\newblock \showarticletitle{The psychology and biology of emotion}.
\newblock


\bibitem[\protect\citeauthoryear{Radford, Wu, Child, Luan, Amodei, and
  Sutskever}{Radford et~al\mbox{.}}{2019}]%
        {radford2019language}
\bibfield{author}{\bibinfo{person}{Alec Radford}, \bibinfo{person}{Jeff Wu},
  \bibinfo{person}{Rewon Child}, \bibinfo{person}{David Luan},
  \bibinfo{person}{Dario Amodei}, {and} \bibinfo{person}{Ilya Sutskever}.}
  \bibinfo{year}{2019}\natexlab{}.
\newblock \showarticletitle{Language Models are Unsupervised Multitask
  Learners}.
\newblock  (\bibinfo{year}{2019}).
\newblock


\bibitem[\protect\citeauthoryear{Ren, Liang, Li, Wang, and de~Rijke}{Ren
  et~al\mbox{.}}{2017}]%
        {ren2017social}
\bibfield{author}{\bibinfo{person}{Zhaochun Ren}, \bibinfo{person}{Shangsong
  Liang}, \bibinfo{person}{Piji Li}, \bibinfo{person}{Shuaiqiang Wang}, {and}
  \bibinfo{person}{Maarten de Rijke}.} \bibinfo{year}{2017}\natexlab{}.
\newblock \showarticletitle{Social collaborative viewpoint regression with
  explainable recommendations}. In \bibinfo{booktitle}{\emph{Proceedings of the
  tenth ACM international conference on web search and data mining}}. ACM,
  \bibinfo{pages}{485--494}.
\newblock


\bibitem[\protect\citeauthoryear{Rorty}{Rorty}{1978}]%
        {rorty1978explaining}
\bibfield{author}{\bibinfo{person}{Am{\'e}lie~Oksenberg Rorty}.}
  \bibinfo{year}{1978}\natexlab{}.
\newblock \showarticletitle{Explaining emotions}.
\newblock \bibinfo{journal}{\emph{The journal of philosophy}}
  \bibinfo{volume}{75}, \bibinfo{number}{3} (\bibinfo{year}{1978}),
  \bibinfo{pages}{139--161}.
\newblock


\bibitem[\protect\citeauthoryear{Sagawa*, Koh*, Hashimoto, and Liang}{Sagawa*
  et~al\mbox{.}}{2020}]%
        {Sagawa*2020Distributionally}
\bibfield{author}{\bibinfo{person}{Shiori Sagawa*}, \bibinfo{person}{Pang~Wei
  Koh*}, \bibinfo{person}{Tatsunori~B. Hashimoto}, {and} \bibinfo{person}{Percy
  Liang}.} \bibinfo{year}{2020}\natexlab{}.
\newblock \showarticletitle{Distributionally Robust Neural Networks}. In
  \bibinfo{booktitle}{\emph{International Conference on Learning
  Representations}}.
\newblock
\urldef\tempurl%
\url{https://openreview.net/forum?id=ryxGuJrFvS}
\showURL{%
\tempurl}


\bibitem[\protect\citeauthoryear{Singh and Joachims}{Singh and
  Joachims}{2018}]%
        {singh2018fairness}
\bibfield{author}{\bibinfo{person}{Ashudeep Singh} {and}
  \bibinfo{person}{Thorsten Joachims}.} \bibinfo{year}{2018}\natexlab{}.
\newblock \showarticletitle{Fairness of Exposure in Rankings}. In
  \bibinfo{booktitle}{\emph{Proceedings of the 24th ACM SIGKDD International
  Conference on Knowledge Discovery and Data Mining}} (London, United Kingdom).
  \bibinfo{publisher}{ACM}, \bibinfo{pages}{2219--2228}.
\newblock
\showISBNx{978-1-4503-5552-0}


\bibitem[\protect\citeauthoryear{Vaswani, Shazeer, Parmar, Uszkoreit, Jones,
  Gomez, Kaiser, and Polosukhin}{Vaswani et~al\mbox{.}}{2017}]%
        {Vaswani}
\bibfield{author}{\bibinfo{person}{Ashish Vaswani}, \bibinfo{person}{Noam
  Shazeer}, \bibinfo{person}{Niki Parmar}, \bibinfo{person}{Jakob Uszkoreit},
  \bibinfo{person}{Llion Jones}, \bibinfo{person}{Aidan~N Gomez},
  \bibinfo{person}{{\L}ukasz Kaiser}, {and} \bibinfo{person}{Illia
  Polosukhin}.} \bibinfo{year}{2017}\natexlab{}.
\newblock \showarticletitle{Attention is all you need}. In
  \bibinfo{booktitle}{\emph{Advances in neural information processing
  systems}}. \bibinfo{pages}{5998--6008}.
\newblock


\bibitem[\protect\citeauthoryear{Yaghoobzadeh, Mehri, Tachet~des Combes, Hazen,
  and Sordoni}{Yaghoobzadeh et~al\mbox{.}}{2021}]%
        {yaghoobzadeh-etal-2021-increasing}
\bibfield{author}{\bibinfo{person}{Yadollah Yaghoobzadeh},
  \bibinfo{person}{Soroush Mehri}, \bibinfo{person}{Remi Tachet~des Combes},
  \bibinfo{person}{T.~J. Hazen}, {and} \bibinfo{person}{Alessandro Sordoni}.}
  \bibinfo{year}{2021}\natexlab{}.
\newblock \showarticletitle{Increasing Robustness to Spurious Correlations
  using Forgettable Examples}. In \bibinfo{booktitle}{\emph{Proceedings of the
  16th Conference of the European Chapter of the Association for Computational
  Linguistics: Main Volume}}. \bibinfo{publisher}{Association for Computational
  Linguistics}, \bibinfo{address}{Online}, \bibinfo{pages}{3319--3332}.
\newblock
\urldef\tempurl%
\url{https://doi.org/10.18653/v1/2021.eacl-main.291}
\showDOI{\tempurl}


\bibitem[\protect\citeauthoryear{Zhang and Chen}{Zhang and Chen}{2020}]%
        {Zhang2018a}
\bibfield{author}{\bibinfo{person}{Yongfeng Zhang} {and} \bibinfo{person}{Xu
  Chen}.} \bibinfo{year}{2020}\natexlab{}.
\newblock \showarticletitle{{Explainable recommendation: A survey and new
  perspectives}}.
\newblock \bibinfo{journal}{\emph{Foundations and Trends in Information
  Retrieval}} (\bibinfo{year}{2020}).
\newblock
\urldef\tempurl%
\url{https://doi.org/10.1561/9781680836592}
\showDOI{\tempurl}


\bibitem[\protect\citeauthoryear{Zhang, Lai, Zhang, Zhang, Liu, and Ma}{Zhang
  et~al\mbox{.}}{2014}]%
        {zhang2014explicit}
\bibfield{author}{\bibinfo{person}{Yongfeng Zhang}, \bibinfo{person}{Guokun
  Lai}, \bibinfo{person}{Min Zhang}, \bibinfo{person}{Yi Zhang},
  \bibinfo{person}{Yiqun Liu}, {and} \bibinfo{person}{Shaoping Ma}.}
  \bibinfo{year}{2014}\natexlab{}.
\newblock \showarticletitle{Explicit factor models for explainable
  recommendation based on phrase-level sentiment analysis}. In
  \bibinfo{booktitle}{\emph{Proceedings of the 37th international ACM SIGIR
  conference on Research \& development in information retrieval}}. ACM,
  \bibinfo{pages}{83--92}.
\newblock


\bibitem[\protect\citeauthoryear{Zhou, Huang, Zhang, Zhu, and Liu}{Zhou
  et~al\mbox{.}}{2018}]%
        {zhou2018emotional}
\bibfield{author}{\bibinfo{person}{Hao Zhou}, \bibinfo{person}{Minlie Huang},
  \bibinfo{person}{Tianyang Zhang}, \bibinfo{person}{Xiaoyan Zhu}, {and}
  \bibinfo{person}{Bing Liu}.} \bibinfo{year}{2018}\natexlab{}.
\newblock \bibinfo{title}{Emotional Chatting Machine: Emotional Conversation
  Generation with Internal and External Memory}.
\newblock
\newblock
\showeprint[arxiv]{1704.01074}~[cs.CL]


\end{thebibliography}

\end{document}